\documentclass[journal]{IEEEtran}
\usepackage{times}  
\usepackage{helvet} 
\usepackage{courier}  
\usepackage[hyphens]{url}  
\usepackage{graphicx} 
\usepackage{caption} 
\DeclareCaptionStyle{ruled}{labelfont=normalfont,labelsep=colon,strut=off} 
\usepackage{xcolor}         
\usepackage{color}         

\usepackage{amsmath}
\usepackage{amsfonts}       
\usepackage{dsfont}
\usepackage{balance}

\usepackage{amsfonts}       
\usepackage{nicefrac}       
\usepackage{microtype}      
\usepackage{xcolor}         
\usepackage{comment}
\usepackage{times}
\usepackage{epsfig}
\usepackage{graphicx}
\usepackage{makecell}
\usepackage{amsmath}
\usepackage{amssymb}
\usepackage[utf8]{inputenc}
\usepackage{ulem}
\usepackage{graphicx}
\usepackage{amsmath}
\usepackage{color}
\graphicspath{ {./Figures/} }
\usepackage{lipsum}
\usepackage{etoolbox}
\let\tinymatrix\smallmatrix

\patchcmd{\tinymatrix}{\scriptstyle}{\scriptscriptstyle}{}{}
\patchcmd{\tinymatrix}{\scriptstyle}{\scriptscriptstyle}{}{}
\patchcmd{\tinymatrix}{\vcenter}{\vtop}{}{}
\patchcmd{\tinymatrix}{\bgroup}{\bgroup\scriptsize}{}{}
\usepackage{algorithmic}
\usepackage[ruled,vlined]{algorithm2e}
\usepackage{listings}
\usepackage[utf8]{inputenc}
\usepackage[T1]{fontenc}
\usepackage{float}
\usepackage{wrapfig}
\usepackage{sidecap}
\usepackage{subcaption,siunitx,booktabs}
\usepackage{placeins}
\usepackage{pifont}
\usepackage[utf8]{inputenc}
\usepackage{colortbl}
\usepackage{dashrule}
\usepackage{ehhline}
\usepackage{arydshln}
\usepackage{mathtools}
\usepackage{cite}

\definecolor{mypink}{rgb}{0.858, 0.188, 0.50}
\definecolor{mygreen}{rgb}{0.16, 0.50, 0.10}
\definecolor{ygreen}{rgb}{0.2, 0.60, 0.20}

\newcommand\note[1]{\textcolor{red}{#1}}

\newcommand\greenps[1]{\textcolor{mygreen}{#1}}
\newcommand{\algosize}{\fontsize{6.9pt}{9pt}\selectfont}
\newcommand{\cmark}{\ding{51}}%
\newcommand{\xmark}{\ding{55}}%
\newcommand\itabold[1]{\underline{{{#1}}}}
\definecolor{codegreen}{rgb}{0.26,0.5,0.5}
\definecolor{codegray}{rgb}{0.5,0.5,0.5}
\definecolor{codepurple}{rgb}{0.58,0,0.82}
\definecolor{weborange}{RGB}{255,165,0}

\lstdefinestyle{mystyle}{
  commentstyle=\color{codegreen},
  keywordstyle=\color{magenta},
  numberstyle=\tiny\color{codegray},    
  stringstyle=\color{codepurple},
  basicstyle=\ttfamily\algosize,
  breakatwhitespace=false,         
  breaklines=true,                 
  captionpos=b,                    
  keepspaces=false, 
  showspaces=false,                
  showstringspaces=false,
  showtabs=false,                  
  tabsize=2,
}

\def \emph{\textit}
\def\eg{\emph{e.g. }}

\def\ie{\emph{i.e., }}

\def\bq{\mathbf{q}}

\def\bP{\mathbf{P}}
\def\bK{\mathbf{K}}
\def\bk{\mathbf{k}}

\def\bP{\mathbf{P}}
\def\bM{\mathbf{M}}
\def\bm{\mathbf{m}}
\def\bx{\mathbf{x}}
\def\bX{\mathbf{X}}
\def\by{\mathbf{y}}
\def\bY{\mathbf{Y}}

\usepackage{soul}
\usepackage[ruled]{algorithm2e}
\usepackage{amssymb,amsmath}
\usepackage[hang,flushmargin]{footmisc}
\usepackage{lipsum}
\makeatletter
\newcommand{\algorithmfootnote}[2][\footnotesize]{%
  \let\old@algocf@finish\@algocf@finish
  \def\@algocf@finish{\old@algocf@finish
    \leavevmode\rlap{\begin{minipage}{\linewidth}
    #1#2
    \end{minipage}}%
  }%
}
\makeatother

\begin{document}
\title{Extending Momentum Contrast with\\ Cross Similarity Consistency Regularization}
\author{Mehdi Seyfi,~\IEEEmembership{Member,~IEEE,}
        Amin Banitalebi-Dehkordi,~\IEEEmembership{Member,~IEEE,}
        Yong Zhang,~\IEEEmembership{Member,~IEEE.}
\thanks{The authors are with the Vancouver Big Data and Intelligence platform Lab, Huawei Technologies Canada Co. Ltd., Vancouver, BC, Canada.}
\thanks{Manuscript received December, 2021.}
\thanks{Copyright © 2022 IEEE. Personal use of this material is permitted. However, permission to use this material for any other purposes must be obtained from the IEEE by sending an email to pubs-permissions@ieee.org.}}

\markboth{IEEE Transactions on Circuits and Systems for Video Technology}{Seyfi \MakeLowercase{\textit{et al.}}: Extending Momentum Contrast with Cross Similarity Consistency Regularization}

\maketitle

\begin{abstract}
Contrastive self-supervised representation learning methods maximize the similarity between the \emph{positive pairs}, and at the same time tend to minimize the similarity between the \emph{negative pairs}.
However, in general the interplay between the negative pairs is ignored as they do not put in place special mechanisms to treat negative pairs differently according to their specific differences and similarities. 
In this paper, we present Extended Momentum Contrast (XMoCo), a self-supervised representation learning method founded upon the legacy of the momentum-encoder unit proposed in the MoCo family configurations \cite{he2019moco, chen2020mocov2, chen2021empirical}. To this end, we introduce a cross consistency regularization loss, with which we extend the transformation consistency to dissimilar images (negative pairs). Under the cross consistency regularization rule, we argue that semantic representations associated with \textit{any} pair of images (positive or negative) should preserve their cross-similarity under pretext transformations.  Moreover, we further regularize the training loss by enforcing a uniform distribution of similarity over the negative pairs across a batch. The proposed regularization can easily be added to existing self-supervised learning algorithms in a plug-and-play fashion.
Empirically, we report a competitive performance on the standard Imagenet-1K linear head classification benchmark.
In addition, by transferring the learned representations to common downstream tasks, we show that using XMoCo with the prevalently utilized augmentations can lead to improvements in the performance of such tasks.
We hope the findings of this paper serve as a motivation for researchers to take into consideration the important interplay among the negative examples in self-supervised learning.
\end{abstract}

\IEEEpeerreviewmaketitle

\section{Introduction}\label{sec_introduction}
The dramatic performance of deep neural networks strongly depends on the amount of training data available. Sophisticated architectures, trained with large scale datasets continue to set new state-of-the-art (SOTA) records. However, collection and annotation of large datasets are time consuming and expensive.
To overcome the costly and laborious data annotation, many self-supervised/unsupervised representation learning methods have been developed to extract visual semantics of the large-scale image datasets, without the burden of labeling them \cite{jing2020self,schmarje2021survey, xmoco_ieee, banitalebi2021repaint, akbari2022lang, banitalebi2021knowledge}. The semantic representations are meant to be preserving useful visual information when representing the images in a downstream task. 

Self-supervised visual representation learning has received a great deal of attention in the recent years, particularly after achieving success in the field of natural language processing \cite{devlin2018bert}.
A popular self-supervised approach to learn the visual representations is to train a network to solve one or multiple pretext tasks on an unlabeled dataset, by introducing objective functions which are optimized through gradient descent. A variety of pretext tasks have been designed so far for self representation learning. Some early, but influential ones include: predicting image rotations \cite{gidaris2018unsupervised}, solving a jigsaw puzzle \cite{noroozi2016unsupervised}, gray scale image colorization \cite{zhang2016colorful}, automatic image inpainting \cite{pathak2016context}, spatio-temporal geometric deformations in a video sequence \cite{xu2020deep}, or utilizing 2D heatmaps as a supervision signal \cite{9123498}.

A breakthrough in designing pretext tasks occurred by the introduction of contrastive learning \cite{Wu_2018_CVPR} and infoNCE loss \cite{oord2018representation,henaff2019data}, 
and was later improved  by the introduction of the \emph{transformation consistency} property
\cite{Misra_2020_CVPR, cvpr19unsupervised}. The authors in  \cite{Misra_2020_CVPR} and \cite{cvpr19unsupervised} reasoned that useful image representations must be invariant under semantics preserving image transformations/augmentations.\footnote{semantic preserving transformations are simple augmentations that do not change the visual semantics of the image, \eg in image classification, image class label is not changed if it undergoes such transformations~\cite{von2021self}.} This observation formed the foundation of the latest generation of contrastive learning algorithms. From the standard contrastive perspective, each image in a batch has one \emph{positive peer} which together they make a \emph{positive pair}, and $K$ \emph{negative peers} which together they make $K$ \emph{negative pairs}. The main idea is to discriminate randomly transformed versions of non-identical images (negative pairs) against each other by minimizing the cosine similarity between their representations. Simultaneously, courtesy of transformation consistency, the features representing transformed versions of identical images (positive pairs) are forced to be close in terms of their normalized inner product \cite{he2019moco,chen2020mocov2,chen2021empirical}. 

The infoNCE loss incorporated in these methods in fact minimizes the cross-entropy between (a) softmax probability of matching a representation to its positive/negative peers, and (b) deterministic one-hot encoded pseudo-labels which always assign 1 to the positive peer and 0 to the negative ones.
Accordingly, the infoNCE loss forces the \emph{cross similarity} between the negative pairs towards zero\footnote{We use the term \emph{cross similarity} for the cosine similarity between the representations associated to negative pairs, since they belong to non-identical images.}. By doing so, infoNCE ignores the information concealed in the cross similarities. However, the negative pairs might be sharing visual semantics with each other. For example, there may be different instances of a same object category. 

To overcome this issue, in this paper, we moderate the effect of infoNCE loss by incorporating the information concealed in the negative samples. To this end, first we propose the idea of \emph{cross similarity consistency regularization}. In particular, we argue that after applying transformations\footnote{In this paper we assume that all the transformations are semantic preserving.}, not only the semantic representation of an image is preserved, but also its cross similarity with the representation associated to each negative peer is preserved. In other words, the representation associated to any random transformation of \emph{image A} is equally similar with the representation of any random transformation of \emph{image B} in the cosine similarity sense. We will later show that using this property we can regularize how the similarity score is distributed between the negative peers.

In addition, we further regularize the learning by softening the pretext pseudo-labels associated to similarities of positive and negative pairs (in the vanilla contrastive approach, this is a one-hot encoded vector). To this end, we generate pseudo-labels from data representation vectors such that in a batch of training data, the probability of examples being similar to the positive peers is encouraged to be high, while their similarity to the negative peers are equally discouraged on the average. The non-zero soft pseudo-label assignments also account for a degree of similarity that exists within the negative pairs. In other words, label smoothing allows the model to learn not only the class with highest prediction score, but also the second most likely prediction, the third, and so on, since it provides relative information across the categories. This is achieved through a constrained entropy maximization technique which we explain later in the paper. This regularization provides extra support to better learn to distinguish between the positive and negative samples, which in turn enhances the representation learning quality.
Our contributions can therefore be summarized as:
\begin{itemize}
    \item We propose a cross similarity consistency regularization strategy to account for the fact that semantic preserving transformations lead to unchanged similarity between representations of negative pairs.  
    \item We provide optimized soft pretext pseudo-labels such that in a batch of data, on the average samples are equally dissimilar to each of the negative peers.\footnote{It must be noted that this technique cannot be applied to algorithms such as \cite{chen2020exploring, caron2020unsupervised} that do not exploit negative samples.}
     \item Our results show competitive performance in the standard linear head classification benchmark on Imagenet-1K and other downstream tasks such as object detection and semantic segmentation.
\end{itemize}
\section{Related Works}
\label{sec:related_works}
This section briefly reviews the related works. Related to our work are in general the self-supervised learning algorithms, and in particular the contrastive approaches.

\subsection{Earlier self-supervised learning techniques}
Initial efforts in self-supervised learning were inspired by applying tangible image operations to extract some of kind of pseudo ground truth to use for training. To this end, \cite{gidaris2018unsupervised} applied a fixed set of rotations to input images and defined a task of rotation classification. They argued that this task will help the model learn generic representations of the images. \cite{noroozi2016unsupervised} defined a pretext task of solving an artificially created jigsaw puzzle. \cite{zhang2016colorful} converted colorful images to gray scale, and designed a colorization auxiliary task. \cite{pathak2016context} used the inpainting of deliberately removed patches to learn representations. A detailed survey of such methods is available in \cite{jing2020self}.
Despite an initial success, the performance of these methods did not reach the levels achieved by contrastive-based methods proposed later. However, they paved the way for some of the strong successors as they showed there is promise in self-supervised learning.
\subsection{Clustering-based Approaches}
\emph{Clustering} methods ratiocinate that data points corresponding to representations belonging to a same cluster must share similar visual semantics. Therefore, they try to minimize the distance between the distributions of the representations generated by the network and cluster assignments of the clustering algorithm \cite{caron2018deep, caron2019unsupervised,caron2020unsupervised,asano2019self,9610083}.

\subsection{Contrastive Learning}
The rationale behind the \emph{contrastive} methods is that different views of a same image, or different frames of the same short video sequence, must be encoded to a same representation. On the other hand, views associated to different images should delegate distant representations in the latent space. Based on this, in their loss they try to drag the representations of the views associated to a same image (positive pairs) towards each other while at the same time push away the representations associated to the views of different images (negative pairs) 
\cite{Wu_2018_CVPR,he2019moco,chen2020mocov2,oord2018representation,Misra_2020_CVPR,9674754,dave2021tclr,9541377}.
\subsection{Siamese Networks}
The \emph{Siamese} approaches eliminate the necessity of incorporating negative samples in their methodology. They argue that the representations of different views of same images should be minimally distant in the representation space. Based on this criteria, they minimize the Euclidean distance between the representations associated to views of same images \cite{chen2020exploring, grill2020bootstrap}.
\subsection{Summary}
In this paper, we extend the existing contrastive learning approach with cross similarity consistency regularization. To this end, we design a contrastive formulation in which not only positive example pairs attain a similar representation, but also negative pairs are equally dis-similar across various semantic preserving transformations. Additionally, we enforce a uniform similarity distribution on the negative examples. In other words, samples across a batch are forced to be on the average, equally similar to each negative peer. These constraints enable the network to incorporate the information concealed in the negative pairs while learning the visual semantics of the dataset.

\section{Method}\label{sec_method}
\begin{SCfigure*}
\centering
\includegraphics[width=.5\textwidth]{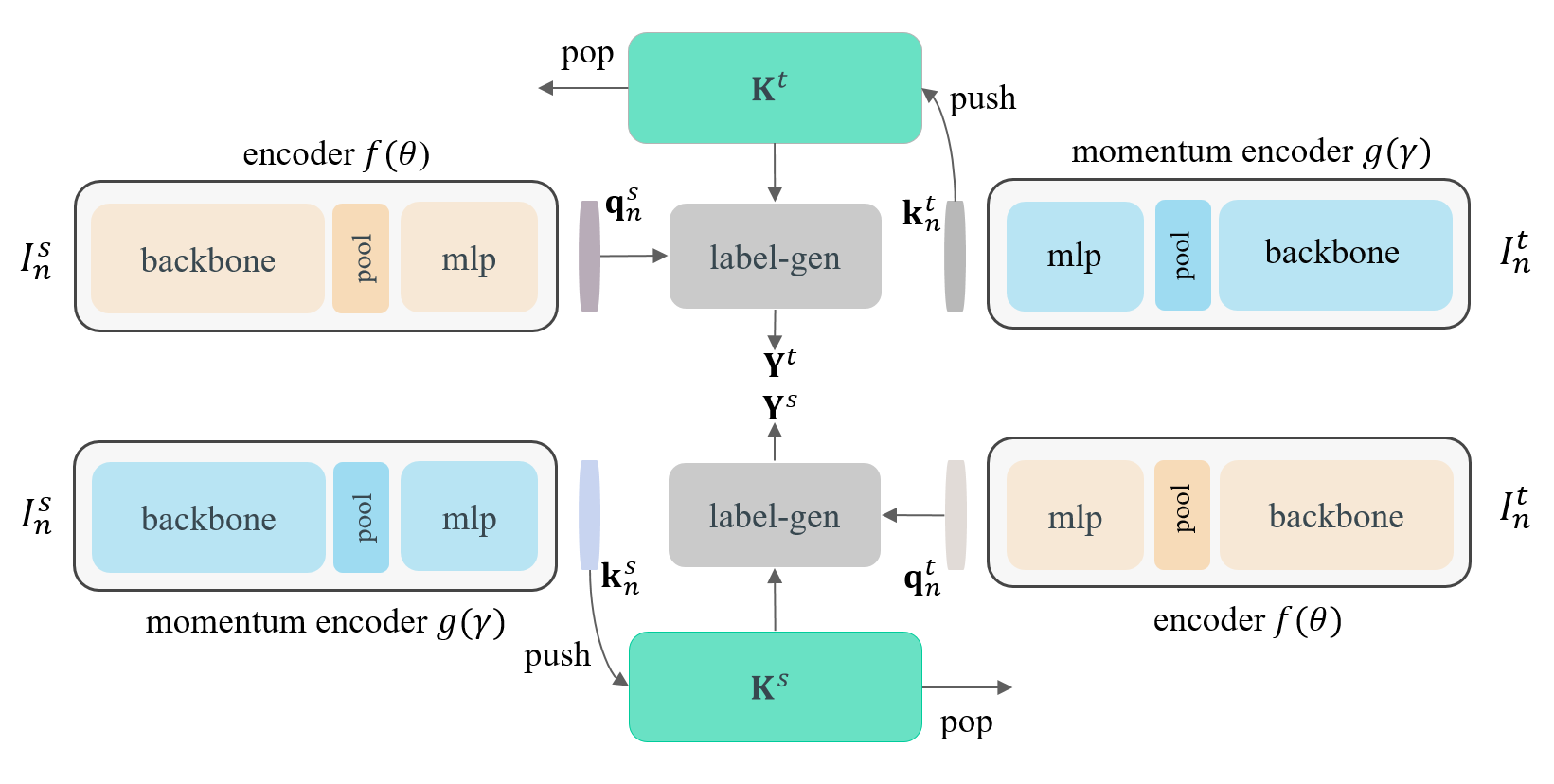}
\caption{The architecture of XMoCo. $I_n^s$ and $I_n^t$ are two views of the same image $I_n$ undergone two random semantic preserving transformations $\mathcal{T}_s$ and $\mathcal{T}_t$, respectively. $I_n^s$ and $I_n^t$ are fed to the encoder and momentum-encoder architectures, resulting in the query and key representation vectors $q_n^s, k_n^t$. The inputs to the encoder and momentum-encoder architectures are then swapped, resulting a new pair of representation vectors $k_n^s, q_n^t$. Based on the representations stored in the two memory banks and the similarities between the representation vectors, pseudo-labels are generated to incorporate cross similarities between the negative pairs as well.}
\label{Figure:XMoCo}
\end{SCfigure*}

This section describes our extended momentum contrast method, dubbed as XMoCo. As a whole, Figure~\ref{Figure:XMoCo} shows the building blocks of our approach. 

Representation vectors/features are obtained by passing images through feature generator networks, and pseudo-labels are created by a pair of label generator functions. The feature generators are two deep convolutional networks, encoder and momentum-encoder, the weights of which are updated by minimizing a predefined loss function. 

The key system components and mechanisms can be summarized as follows:
\begin{enumerate}
    \item Each image in a batch is transformed into two versions which are called \emph{source} and \emph{target} views. These views are generated by passing the images through random semantic preserving transformations. Subsequently, the source and target \emph{features} are the embeddings inferred by passing the their associated views through the network. For $N$ images in the batch we have total of $2N$ views.
    \item Since the images are passed through \emph{semantic preserving transformations}, their visual meaning inferred by looking at transformed image is not altered.
    \item For each source view there exists \emph{one, and only one,} view among the $N$ target views which is co-generated from the exact same image and vice versa. These two views form a \emph{positive pair}.
    \item For each source view there exists exactly $N - 1$ views (\emph{negative peers}) from different original images in the batch, and vice versa. Together, they constitute $N - 1$ \emph{negative pairs} of views. 
    \item In each batch, $N - 1$ source and $N - 1$ target features are stored in two long memory banks.
    \item The source features should be similar to their associated target features because the have the same origin.
    \item Each source feature should be dissimilar (to a certain degree) to all the negative features stored in the target memory bank, because they are not from same origin. Similarly, each target feature should be dissimilar (to a certain degree) to all the stored negative features in the source memory bank for the same reason. 
\end{enumerate}

Next, we provide basic definitions and setup of the problem, followed by our solution description.

\subsection{System Model}
Let $\mathcal{I} = \{I_1, \ldots, I_{\mathcal{|I|}}\}$ denote a training dataset comprising $\mathcal{|I|}$ number of images. Also, assume $\mathcal{T}_s(\cdot)$ and $\mathcal{T}_t(\cdot)$ are two image transformation functions,
sampled randomly from a continuous set of semantic preserving transformations $\{\mathcal{T}_i(\cdot)\}$ formed by a sequential composition of random crop and resize, color jitter, random gray scale, and random horizontal flip \cite{tian2020makes,grill2020bootstrap,caron2020unsupervised,Misra_2020_CVPR,chen2020simple,chen2020big}. 

Thus, for the $n^{th}$ image in a batch denoted by $I_n$, we can generate two different transformed versions $I_n^s = \mathcal{T}_s(I_n)$ and $I_n^t = \mathcal{T}_t(I_n)$, respectively. We denote $I_n^s$ as the \emph{source} and $I_n^t$ as the \emph{target} view, the representations of which are later forced to be transformation consistent. Following the general momentum contrast strategy of \cite{he2019moco, chen2020mocov2, chen2021empirical}, for $\mathcal{T} \in \left\{\mathcal{T}_i(\cdot)\right\}$, we use an encoder network $f:\mathcal{T}(I_n)\longrightarrow \mathds{R}^d$, which consists of a backbone convolutional network, a pooling layer, and a set of sequential fully connected linear layers followed by a final normalization layer\footnote{For simplicity of notation we assume the encoder and the momentum-encoder architectures provide normalized feature vectors.}. Moreover, as shown in Figure \ref{Figure:XMoCo}, the momentum-encoder network $g:\mathcal{T}(I_n):\longrightarrow \mathds{R}^d$ has a similar structure. The encoder network generates the \textit{query} representation vector $\bq_n = f(\mathcal{T}(I_n))$. Simultaneously, the momentum-encoder network generates the \textit{key} representation vector $\bk_n = g(\mathcal{T}(I_n))$. 

Our system design also contains two pre-registered memory banks of size $d\times K$, where $K$ is the memory bank length. After each iteration of training, the memory banks are updated by the key feature vectors in the batch. These banks are updated via the same mechanism as in \cite{he2019moco, chen2020mocov2, chen2021empirical}. 

In our architecture, the query vectors $\bq_n^s$ and $\bq_n^t$, are obtained by passing the source and the target views of image $I_n$, \ie $I_n^s$ and $I_n^t$, through the encoder $f(\cdot)$. 
Following the standard practice in the contrastive learning literature, we choose a Resnet50 encoder with its fully connected layer, $\texttt{res50.fc}$, replaced with a multi layer perceptron \cite{grill2020bootstrap,caron2020unsupervised,Misra_2020_CVPR,chen2020simple,chen2020big}. At the same time, $I_n^s$ and $I_n^t$, traverse through the momentum-encoder structure $g(\cdot)$ and create the key feature vectors $\bk_n^s$ and $\bk_n^t$, respectively. Therefore, $\bq_n^s = f(I_n^s)$, $\bq_n^t = f(I_n^t)$, $\bk_n^s = g(I_n^s)$, and $\bk_n^t = g(I_n^t)$ are different representations of $I_n^s$ and $I_n^t$, via separate networks.
Since the images are not repeated in the batches during one epoch, it is guaranteed that there is no representation in the memory bank that is similar to any representations in the current batch. Therefore we can safely use all the representations in the memory as negative features.

The two memory banks denoted by $\bK^s\in\mathds{R}^{d\times K}$and $\bK^t\in\mathds{R}^{d\times K}$, keep track of the source and target key features from \emph{previous} iterations. For ease of notation, we define  $\bM^s = \left[\bm_1^s, \ldots, \bm_{K+1}^s\right]$ belonging to $\mathds{R}^{d\times (K+1)}$, where $\left[\bm_2^s, \ldots, \bm_{K+1}^s\right]=\bK^s$, are the stashed source key features pertaining to previous iterations, and $\bm_1^s = \bk_n^s$ holds the \emph{current} source feature. In a similar way $\bM^t = \left[\bm_1^t, \ldots, \bm_{K+1}^t\right] \in \mathds{R}^{d\times (K+1)}$ holds the current target key, $\bm_1^t = \bk_n^t$ and the set of target key representations $\left[\bm_2^t, \ldots, \bm_{K+1}^t\right]$ belonging to the previous iterations,  such that $\left[\bm_2^t, \ldots, \bm_{K+1}^t\right]=\bK^t$.

Throughout this paper unless otherwise mentioned we use $\mathcal{L}$ to notify the average cross entropy loss of distribution $\bX_2$ relative to distribution $\bX_1$ in a batch of $N$ samples via
\begin{equation}\label{eq_general_l}
\mathcal{L}(\bX_1,\bX_2)= \min_{\bX_1}~\frac1N\mathds{E}_{_{\mathcal{T}}}\left\{ \left<\bX1, -\log \bX_2\right> \right\},
\end{equation}
where $\left<\cdot\right>$ is the Frobenius dot product between probability matrices $\bX_1$ and $\bX_2$. 

\subsection{ Classical Contrastive Learning}\label{sec_cne}
The idea of contrastive learning is to homogenize the positive pairs, ($\bq_n^s$, $\bm_1^t)$ and $(\bq_n^t, \bm_1^s)$, by forcing them to be similar in the feature space via a cosine similarity function:
$$
S (\bx, \by)= \frac{\bx^T\by}{\|\bx\|_2\|\by\|_2}.
$$
Simultaneously, the representation extracted from image $I_n \in \mathcal{I}$ is contrasted with the representations associated to other images in $\mathcal{I}^-\subseteq\mathcal{I}\backslash \{I_n\}$ with $|\mathcal{I}^-|=K$. 
In other words, for $k= 2, \ldots, K+1$, traditional contrastive learning is interested in making the negative pairs $(\bq_n^s, \bm_k^t)$ and $(\bq_n^t, \bm^s_k)$  as dis-similar as possible.

In fact $S(\bq_n^s,\bm_k^t)$ is the distance between the query representation $\bq_n^s$, and the $k^{th}$ negative peer representation vector. Based on the relative distance between $\bq_n^s$ and all other negative peer representation vectors, $\bm_k^t$, the posterior softmax probability of the query feature $\bq_n^s$ being correlated/similar to the representation vector associated to the peer $k$ is given by:
\begin{eqnarray}\label{eq_cne_I}
p(\bm_k^t,\bq_n^s|I_n)=\qquad\qquad\qquad\qquad\quad\qquad\qquad\qquad&&\nonumber\\ \frac{\exp\left(\frac{S\left( \bq_n^s,\bm_k^t\right)}{\tau}\right)}{\exp\left(\frac{S\left(  \bq_{n}^s, \bm_k^t\right)}{\tau}\right) + \sum\limits_{l\neq k} \exp\left(\frac{S\left(\bq_{n}^s,  \bm_l^t\right)}{\tau}\right)},&&
\end{eqnarray}
where the peer is positive for $k=1$ and negative otherwise. It must be emphasised that $\bm^t_1$ represents $ I^t_n$, and the rest of the key features are taken from $\bK^t$.
The temperature $\tau$, is a hyper-parameter value that must be fine-tuned.
In a similar way, the posterior probability that the query feature $\bq_n^t$ be similar to the representation vector related to the $k^{th}$ peer is given by:
\begin{eqnarray}\label{eq_cne_II}
p(\bm_k^s,\bq_n^t|I_n)=\qquad\qquad\qquad\qquad\quad\qquad\qquad\qquad&&\nonumber\\ \frac{ \exp\left(\frac{S\left( \bq_{n}^t, \bm_k^s\right)}{\tau}\right)}{\exp\left(\frac{S\left(  \bq_{n}^t, \bm_k^s\right)}{\tau}\right) + \sum\limits_{l\neq k} \exp\left(\frac{S\left(  \bq_{n}^t,  \bm_l^s\right)}{\tau}\right)},&&
\end{eqnarray}
where $\bm^s_1$ represents $I_n^s$.
Note that all the features in the memory banks $\bK^s$ and $\bK^t$ are generated by the momentum-encoder architecture, the weights of which are frozen and are only updated via a momentum update recursion \cite{he2019moco}.

In the classical contrastive learning, a one-hot pseudo-label $y_k\in\{0,1\}$ is assigned to every $K+1$ peers with the deterministic posterior probability distribution:
\begin{equation}\label{eq_prob_dist_pseudo_labels}
p(y_k|I_n) = \delta(y_k-y_1),\qquad ~k\in[1,K+1].
\end{equation}
The \emph{symmetric} contrastive scheme \cite{Misra_2020_CVPR, chen2021empirical}, therefore minimizes the average of cross-entropy losses between the probabilities in \eqref{eq_cne_I} and \eqref{eq_cne_II} and the pseudo-labels $y_k$ via
\begin{eqnarray}
\mathcal{L} = -\frac1{N}\sum_{n=1}^N\left(\sum_{k=1}^{K+1} p(y_k|I_n) \log(p(\bm_k^t,\bq_n^s|I_n))\right.&&\nonumber\\
\left.+ \sum_{k=1}^{K+1} p(y_k|I_n) \log(p(\bm_k^s,\bq_n^t|I_n))\right),&&\nonumber\\
=\frac{-\sum_{n=1}^N\log(p(\bm_1^t,\bq_n^s|I_n)) +\log(p(\bm_1^s,\bq_n^t|I_n))
}{N}.&&
\end{eqnarray}
\subsection{Cross Similarity Consistency}\label{sec_cross_similarity_consistency}
Our idea of cross similarity consistency is declared as follows: For any pair of transformations $(\mathcal{T}_s, \mathcal{T}_t)\in \{\mathcal{T}_i(.)\}$, and $\{(I_n, I_k)|I_n\in\mathcal{I}, I_k \in\mathcal{I}^-, k=2, \ldots, K+1\}$:
\begin{equation}\label{eq_cross_sim}
S(\bq_n^s, \bm_k^t) = S(\bq_n^t, \bm_k^s). 
\end{equation}
The cross similarity consistency in \eqref{eq_cross_sim} suggests that under any semantic preserving transformation, the similarity of \emph{negative pairs} remains unchanged. This in turn stems from the transformation invariance hypothesis; meaning that, for any image $I_n\in\mathcal{I}$ the representation vector associated to the transformed image $\mathcal{T}(I_n)$ remains unchanged for any $\mathcal{T}\in\{\mathcal{T}_i(.)\}$.

Applying the cross similarity consistency in \eqref{eq_cross_sim} to \eqref{eq_cne_I} and \eqref{eq_cne_II} calls for:  
\begin{equation}\label{eq_cross_sim_impose}
p(\bm_k^t,\bq_n^s|I_n) = p(\bm_k^s,\bq_n^t|I_n).
\end{equation}
Note that the probabilities in \eqref{eq_cne_I} and \eqref{eq_cne_II} are in fact \emph{normalized similarities} of positive and negative pairs.
\subsection{Uniform Similarity to Negative Peers}\label{subsec_Uniform Similarity to Negative Peers}
As mentioned earlier, for any image $I_n$ in a batch, the classical contrastive learning assumes a one-hot encoded pretext pseudo-label with probability distribution defined in \eqref{eq_prob_dist_pseudo_labels} which assigns $f(I_n)$ to the positive peer with a deterministic probability of 1. This choice of pseudo-labeling ignores the so called dark knowledge \cite{hinton2015distilling} in the cross similarities between the query and the negative key features, and imposes the similarity of the query feature to any of the negative peers with a zero probability, whereas the soft labels have high entropy, providing much more information per training sample than one-hot-encoded labels~\cite{hinton2015distilling}. We propose to soften the pseudo-labels by considering the outcome of cross similarity consistency in \eqref{eq_cross_sim_impose}. 
\begin{figure*}[htb!]
\centering
\begin{subfigure}[b]{0.45\textwidth}
   \includegraphics[width=1\linewidth]{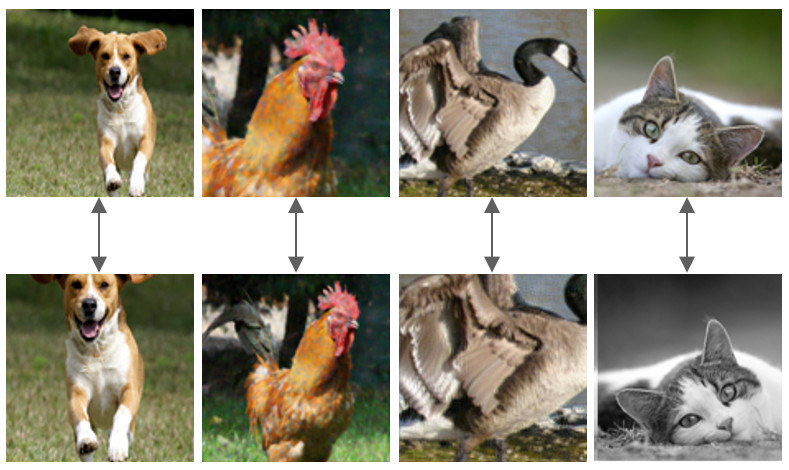}
   \caption{}
   \label{fig:Ng1} 
\end{subfigure} \qquad \qquad
\begin{subfigure}[b]{0.45\textwidth}
   \includegraphics[width=1\linewidth]{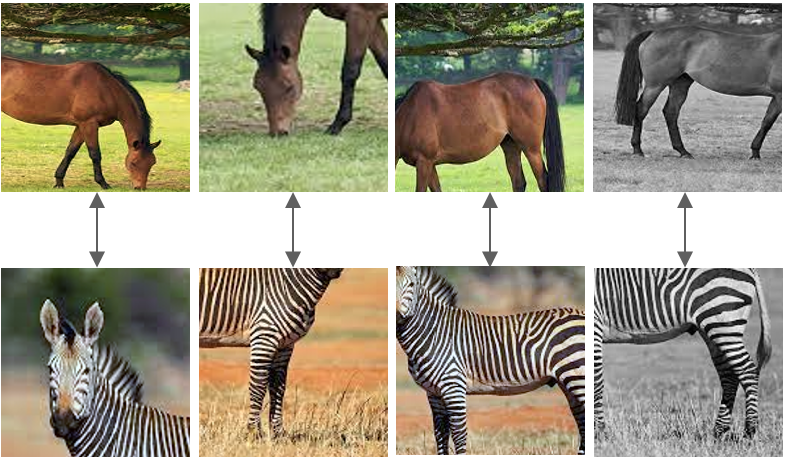}
   \caption{}
   \label{fig:Ng2}
\end{subfigure}
\vspace{-8pt}
\caption[Transformation Consistency]{For any semantic preserving transformation: (a) \textbf{Transformation invariance consistency:} Images in the same columns make positive pairs, respectively. The representations associated to images in each row are similar to the their positive peer counterparts in the other row. Basically, random augmentations, \eg cropping, horizontal flip, etc, have separately been applied to original images to get the first and second row images. The images in the first row contain the same information as their second row counterparts, since the contextual meanings of the images do not change through transformation; therefore, their  associated one-to-one representations are similar in the embedding space. This is what we mean by transformation invariance consistency.}  (b) \textbf{Cross similarity consistency:} is a by-product of transformation invariance consistency. Images in the same columns make negative pairs. They come from images with different contextual semantics. In the traditional contrastive learning, their representation features should not be similar at all (zero cosine similarity). But, in this paper we let the model capture some similarities among the negative pairs (\eg zebra and horse are not the same but they share visual semantics). If the cosine similarity between the images in the first column is $\alpha$, then the similarity between the images in every other column is also $\alpha$.
\label{fig:transformation_invariance_consistency}
\end{figure*}
Let the pseudo-labels associated to $I^s_n$ and $I^t_n$ be defined as $\by^s_n$ and $\by^t_n$, where 
$[\by^s_n]_{_k} = p(y_k|I_n^s)$ and $[\by^t_n]_{_k} = p(y_k|I_n^t)$.  
We denote the pseudo-labels in a batch of $N$ samples by $\bY^s = \left[\by^{s}_{1}, \ldots, \by^s_N\right]$ and $\bY^t = \left[\by^t_1, \ldots, \by^t_N\right]$. Let's also define probability matrices $\bP^s$ and $\bP^t$ in $\mathds{R}^{(K+1)\times N}$ such that
$\left[\bP^s\right]_{_{kn}} = p(\bm_k^t,\bq_n^s|I_n)$, and $\left[\bP^t\right]_{_{kn}} = p(\bm_k^s,\bq_n^t|I_n)$.
We propose to converge the probability distribution of $\bP^s$ to $\bY^t$ and $\bP^t$ to $\bY^s$ under certain conditions:
\begin{itemize}
    \item The probability distribution of the pseudo-labels for the positive peer must be larger than the negative peers, since the positive pairs are more likely to be similar.
    \item On the average, representation vectors associated to images in the batch are equally similar/dis-similar to any of the negative representations in the memory bank. In other words, in a batch of data, on the average, the similarities, are uniformly distributed among all the negative peers. 
\end{itemize}
We train the network by alternating between two optimization problems. First, we find the closest distribution of pseudo-labels to the output probability distributions with the constraints we just mentioned. In particular, we find the optimal pretext pseudo-labels via the objective function:
\begin{equation}\label{eq_pseudo_label}
\mathcal{L}_{pl} = \mathcal{L}_{\mu}(\bY^s,\bP^s) + \mathcal{L}_{\mu}(\bY^t,\bP^t),
\end{equation}
where
\begin{equation}\label{eq_max_entropy}
        \begin{aligned}
            &\mathcal{L}_{\mu}(\bY,\bP)= \min_{\bY}~\frac1N\mathds{E}_{_{\mathcal{T}}}\left\{ \left<\bY, -\log \bP\right> \right\},\\
            \textrm{s.t.}~~ &[\bY]_{kn} \geq 0,\\[-3pt]
            &\sum\nolimits_{k=1}^{K+1} [\bY]_{kn} = 1,~\qquad\qquad\qquad n\in[1, N],\\[-3pt]
            &\sum\nolimits_{n=1}^N [\bY]_{_{1n}} =N\xi, \\[-3pt]
            &\sum\nolimits_{n=1}^N [\bY]_{_{kn}}=\frac{N(1 - \xi)}{K}, \quad~ k\in(1,K+1],\\[-3pt]
             &[\bY]_{_{1n}} \geq \max_k [\bY]_{kn} ,~\qquad\qquad\qquad n\in[1, N],
        \end{aligned}
\end{equation}
for $1/ (K + 1)\leq \xi \leq 1 $.
The first and the second constraints in \eqref{eq_max_entropy} guarantee that the pseudo-labels are indeed probabilities. The third and the last constraints are imposed by the infoNCE transformation invariance consistency and guarantee the positive pair is the most similar pair. Finally, the forth constraint pushes the pseudo-labels to be uniformly distributed over the negative peers in a batch of data. 
It is worth noting that $\xi = 1$ converts the optimization of \eqref{eq_max_entropy} to the classical contrastive learning problem.
To solve \eqref{eq_max_entropy} we relax the last constraint by letting $[\bY]_{1n} = \xi$. This would guarantee that $\bY_{1n} \geq \max [\bY]_{kn}$. Removing the constraint for $[\bY]_{1n}$, our new optimization problem becomes:
\begin{equation} \label{eq_max_entropy_II}
\begin{aligned}
&\mathcal{L}_{\mu}(\bY,\bP)= (1-\xi)\min_{\hat{\bY}}~\mathds{E}_{_{\mathcal{T}}}\left\{ \left<\hat{\bY},  -\log\hat{\bP}\right> \right\} + \mathcal{C}_0 \\
       \textrm{s.t.}~~ &[\hat{\bY}]_{kn} \geq 0,\\[-3pt]
            &\sum\nolimits_{k=1}^{K} [\hat{\bY}]_{kn} = \frac1N,~\quad\qquad\qquad n\in[1, N],\\[-3pt]
            &\sum\nolimits_{n=1}^N [\hat{\bY}]_{_{kn}}=\frac{1}{K},~\quad\qquad\qquad k\in[1,K],
        \end{aligned}
\end{equation}
where $\hat{\bY}$ and $\hat{\bP}$ are sub-matrices of $\bY$ and $\bP$ with their first row removed such that $[\hat{\bY}]_{kn} = \frac{[\bY]_{(k+1)n}}{N(1-\xi)}$, and $[\hat{\bP}]_{kn} = \frac{[\bP]_{(k+1)n}}{N}$,
and $\mathcal{C}_0=-\frac{\xi}{N}\sum_{n=1}^N{\log[\bP]_{1n}}-(1-\xi)\log N$ is a constant. 
Note that $\hat{\bY}\mathds{1} = r$, and  $\hat{\bY}^T\mathds{1} = c$, 
where $\mathds{1}$ is a vector of all ones with proper dimensions, and $r= \frac1K.\mathds{1}$ and $c= \frac1N.\mathds{1}$ are histograms with $r^T\mathds{1}=1$ and $c^T\mathds{1}=1$. 
The optimization problem in \eqref{eq_max_entropy_II} is a transportation problem for which a fast solution is provided by \cite{cuturi2013sinkhorn} by smoothing the cost function with an entropic regularization
term,$-\frac1{\lambda}\mathcal{H}(\hat{\bY})$. 
The author in \cite{cuturi2013sinkhorn} showed that using a fast version of the \emph{Sinkhorn-Knopp} algorithm, the minimizer of  \eqref{eq_max_entropy_II} for a proper value of $\lambda$ can be obtained by: 
\begin{equation}\label{eq_opt_z}
\hat{\bY} = \textrm{diag}(\alpha){\hat{\bP}}^{\lambda}\textrm{diag}(\beta).
\end{equation}
Note that -$\mathcal{H}(\hat{\bY}) =\sum_{kn}[\hat{\bY}]_{_{kn}}\log{[\hat{\bY}]_{_{kn}}}$, is the negative entropy of the pseudo-labels over the negative peers, which filters out the solutions with smaller entropy. The normalization factors $\alpha$ and $\beta$ can be found by simple matrix scaling such that $\hat{\bY}$ satisfies the constraints in \eqref{eq_max_entropy_II}. 
Eventually the optimal pseudo-labels are assigned via\footnote{To be more clear the pseudo labels on each side are obtain from $\bY_s = \begin{bmatrix}
\mathds{1}.\xi & N(1-\xi){\hat{\bY}_s}^T
\end{bmatrix}^T,$ and $\bY_t = \begin{bmatrix}
\mathds{1}.\xi & N(1-\xi){\hat{\bY}_t}^T
\end{bmatrix}^T$, respectively.}: 
\begin{equation}\label{eq_optimal_Y}
\bY = \begin{bmatrix}
\mathds{1}.\xi & N(1-\xi)\hat{\bY}^T
\end{bmatrix}^T.
\end{equation}

Once the pseudo-labels are obtained, we minimize the \emph{symmetric cross-entropy} between the pseudo-labels and the probability distributions and the \emph{cross similarity regularization} cross-entropy via:
\begin{eqnarray}\label{eq_cross_entropy_net}
\mathcal{L}_{ce} &=& \mathcal{L}_{\nu}(\bY^s,\bP^t) + \mathcal{L}_{\nu}(\bY^t,\bP^s)\nonumber \\&+& \underbrace{\mathcal{L}_{\nu}(\bP^s,\bP^t) + \mathcal{L}_{\nu}(\bP^t,\bP^s)}_{\text{cross similarity regularization}}.
\end{eqnarray}  


\begin{algorithm}[htb!]
\SetAlgoLined
\begin{lstlisting}[language=Python]
# f: encoder, g: momentum-encoder
# tau: temperature,  N: batch size
# Ks: memory bank for source view 
# Kt: memory bank for target view
# K: number of negative samples
g.params = f.params # initialize
for x in loader:
    xs, xt = aug(x), aug(x)
    qs, qt = f(xs), f(xt)
    ks, kt = g(xs), g(xt)
    
    Ps = Get_Prob(qs, kt, Kt)
    Pt = Get_Prob(qt, ks, Ks)
    
    # get the loss    
    loss = L(Ps, Pt) 
    
    # SGD update: encoder network
    loss.backward()
    update(f.params)
    # momentum update: momentum-encoder network
    g.params = m * g.params + (1 - m) * f.params
    # update dictionary
    enqueue(ks, Ks) # enqueue the current minibatch
    enqueue(kt, Kt) # enqueue the current minibatch
    dequeue(Ks, Kt) # dequeue the earliest minibatch

def Get_Prob(q,k,K):
    # no gradient to keys
    k = k.detach()
    # positive logits: Nx1
    l_pos = bmm(q.view(N,1,d), k.view(N,d,1))
    # negative logits: NxK
    l_neg = mm(q.view(N,d), K.view(d,K))
    # logits: Nx(1+K)
    logits = cat([l_pos, l_neg], dim=1)/ tau
    # prob. matrix in Eqn.(2),(3)
    P = softmax(logits,dim=1)
    return P

def L(Ps,Pt):
    # find Pseudo Labels from Eq. (12)
    Ys = sinkhorn(Ps.detach())
    Yt = sinkhorn(Pt.detach())
    
    # update the network weights via Eq. (13)
    loss = -((Ys * torch.log(Pt)).sum(dim=1) + 
            Yt * torch.log(Ps)).sum(dim=1) + 
            (Ps.detach() * torch.log(Pt)).sum(dim=1) + 
            (Pt.detach() * torch.log(Ps)).sum(dim=1) +
            ).mean()
    return loss

def sinkhorn(out, lambda, niters=3):#Eq.(11) and Eq.(12)
    N , K = out.shape[0], out.shape[1] - 1
    Y = torch.pow(out[:, 1:] / N ,lambda).t()
    Y /= torch.sum(Y)
    for it in range(niter):
        Y /= torch.sum(Y, dim=1, keepdim=True) * K
        Y /= torch.sum(Y, dim=0, keepdim=True) * N
    Y = torch.cat([xi * torch.ones_like(Y[0, :]).unsqueeze(0), ((1 - xi) * N) * Y], dim=0)# Eq.(12) 
    return Y.t()

\end{lstlisting}
\caption{XMoCo: Pytorch-like Psudeo code}
\label{algo:pseudo_code}
\end{algorithm}

The network is trained by alternating between generating pseudo-labels and updating the network weights.
In this scheme in each iteration two set of pseudo-labels $\bY^s$ and $\bY^t$ are first generated via \eqref{eq_optimal_Y}. Then these pseudo-labels are used to update the networks weight by back-propagating the gradients of loss function in \eqref{eq_cross_entropy_net} through the network using stochastic gradient decent algorithms. Once the weights are updated this procedure repeats. Therefore the network converges by alternating the following two steps:
\begin{enumerate}
\item Generate the optimal pseudo-labels $\bY^s$ and $\bY^t$ are obtained from \eqref{eq_optimal_Y}. 
\item Optimise \eqref{eq_cross_entropy_net} with respect to $\bP^t$ and $\bP^s$ and back-propagate.
\end{enumerate}
Algorithm \ref{algo:pseudo_code} shows our pseudo code in pytorch format.
\subsection{Discussion}\label{sec_Discussion}
Each element of $\bP^s$ and $\bP^t$ is in fact a cross similarity between representations of either a positive or a negative pair. It must be noted that $[\bP^s]_{nk}$ and $[\bP^t]_{nk}$ belong to transformed versions of identical images. 
The first two terms in \eqref{eq_cross_entropy_net} are responsible to uniformly scatter the similarities between the negative pairs; however, the second two terms push the cross similarities to follow \eqref{eq_cross_sim_impose}. Equation \eqref{eq_cross_entropy_net} therefore, minimizes the cross entropy between $\bP^s$ and $\bP^t$ via some pseudo-labels $\bY^t$ and $\bY^s$. Minimizing the average column-by-column cross entropy between $\bP^s$ and $\bP^t$ implies that no matter what the transformation is, as long as it is semantic preserving, the cross similarity between both positive and negative pairs should remain the same. 

Although not mathematically required, it is intuitively feasible to set $N \geq K$.
Throughout this paper we use $N\leq K$, however following \cite{caron2020unsupervised} we use a queue of past probabilities to incorporate in the loss. 

\begingroup
\setlength{\tabcolsep}{6pt} 
\renewcommand{\arraystretch}{1} 
\begin{table*}[tbh!]
\centering
\begin{tabular}{c}
Top-1 accuracy for frozen linear head vs end-to-end fine-tuning\\ 
\hline
\begin{tabular}{lcccccc}
    pre-train&\begin{tabular}{c}batch\\size\end{tabular}&\begin{tabular}{c}200 \\epochs\end{tabular} &\begin{tabular}{c}$\left[800-1000\right]$\\epochs\end{tabular}& \begin{tabular}{c}symmetric\\loss\end{tabular}& \begin{tabular}{c}MLP\\layers\end{tabular}&\begin{tabular}{c}{Fine tune}\\{20 epochs}\end{tabular}\\
    \Xhline{2\arrayrulewidth}
Supervised&&75.9&\\
    \Xhline{2\arrayrulewidth}

SimCLR*\cite{chen2020simple}      & 4096 &68.3     & 70.4         &\cmark & 3&{69.2}\\
MoCo-v2+*\cite{chen2020simple}    & 256   &69.9    & 72.2         &\cmark & 3&{70.5}\\
{MoCo-v2+align/uniform\cite{wang2020understanding}}&{256}&{67.7}&{-}&{\cmark}&{2}&{68.6}\\
BYOL*\cite{grill2020bootstrap}    & 256   &70.6    &\textbf{74.3} & \cmark& 2&{71.3}\\  
SwAV*\cite{caron2020unsupervised} & 4096  &69.1    & 71.8         &\cmark & 3&{70.0}\\
SimSiam\cite{chen2020simple}      & 256   &70.0    & 71.3         &\cmark & 3&{71.2}\\
 {JCL\cite{cai2020joint}}            & {512}      &{68.7}    &{71.0}         &{\xmark} & {2}&{69.5}\\
 {SwAV+UTOA*\cite{wang2021improving}} &{256}&{69.8}&{72.5}       &{\cmark}        &{3}&{70.5}\\
BarlowT\cite{zbontar2021barlow}   & 2048  &\itabold{71.0}    & 73.2         &\xmark & 3&{\textbf{71.9}}\\
{VCReg*\cite{bardes2022vicreg}}   & {256}   & {70.8}   & {73.2}         &{\cmark} & {3}&{71.7} \\
    \hline
XMoCo(ours)                       & 512   & \textbf{71.3}  & \itabold{74.0} &\cmark& 4&{\itabold{71.8}}
\end{tabular}
\end{tabular}
\caption{\footnotesize{Linear head is trained on top of frozen pre-trained backbone. All the backbones are Resnet50$\times 1$ with 25 million parameters. For fair comparison, all the pre-training experiments are performed with two views of size $224 \times 224$ per image. Additionally the methods designated by $*$ are reproduced versions of the original papers with adding symmetry to their loss and/or increasing the number of their MLP layers \cite{chen2020simple}. The second best top-1 accuracy is underlined in each scenario. {The last column depicts the top-1 accuracy performance when the network is fine-tuned for 20 epoch on Imagenet-1K.}}}
\label{tab:linear_head_cls}
\end{table*}
\endgroup

\begin{table}
\centering
\begin{tabular}{lcccc}
     layers&2&3&4&5\\
         \Xhline{2\arrayrulewidth}
     top-1 acc&         69.0&70.6&\textbf{71.3}&70.0\\
     time/200 ep &57h&59h &60h &62h         \\
    memory/64 batch-size&4.5G&4.6G&4.7G&4.8G
\end{tabular}
\caption{Effect of MLP layers in linear head evaluation on Imagenet-1K, when pre-trained for 200 epochs.}
\label{tab:effect_mlp_layers}
\end{table}

\begingroup
\setlength{\tabcolsep}{6pt} 
\renewcommand{\arraystretch}{1.01} 
\begin{table}[htb!]
\centering
\begin{tabular}{c}
\hline
Ablation on the number of MLP layers of other SOTA methods\\ 
\Xhline{2\arrayrulewidth}
\begin{tabular}{lccc}
    Algorithm& Original & 3 layers &4 layers\\
    \hline
    SimCLR*\cite{chen2020exploring,chen2020simple,chen2020improved}& 66.6 (2 layers)& 68.3 &68.0\\
    MoCo-v2+*\cite{chen2020exploring}& 67.5 (2 layers)& 69.9&69.8\\
    BYOL*\cite{chen2020exploring,grill2020bootstrap}&70.6 (2 layers)& 70.3&70.2\\
    SwAV*\cite{caron2020unsupervised,chen2020exploring}& 68.4 (2 layers)&69.1 &69.0\\
    SimSiam*\cite{chen2020exploring}& 71.0 (3 layers)&71.0 &71.1\\
\end{tabular}
\end{tabular}
\caption{Increasing MLP layers does not guarantee learning better features. Training is performed for 200 epochs.}
\label{tab:mlp_other_sota}
\end{table}
\endgroup

\section{Experiment Results}\label{sec:ex-results}
In this section we evaluate the performance of our method and compare 
our results with the existing approaches. Details of experiments can be seen in the Appendix.\\
\textbf{General Setup: }Following the common practice we use a Resnet50 architecture as our backbone. We also use two views of size $224\times 224$ during Imagenet-1K pre-training. In our structure we use a multi layer MLP projection head where each layer consists of a fully connected layer, followed by batchnorm and relu. We use the same \texttt{SGD} optimizer and cosine learning rate scheduler as in \cite{he2019moco} with a base learning rate of $.0675\times \textrm{batch size}/256$ for batch size of 512. For a fair comparison, denoted by $``*"$, the results of other papers are reproduced, either by us or different authors, following the standard practice of two $224\times224$ views for pre-training. Unless otherwise stated, we set $\tau=.2$, $d=128$, $K=4096$ and $\lambda=2$ for all the experiments. The memory bank and the momentum update parameters are adopted from \cite{he2019moco}.
\begin{table}
\centering
\begin{tabular}{lcccccc}
     $\xi$& 0.5  & 0.6 &  0.7  &  0.8  &  0.9  & 1.0\\
         \Xhline{2\arrayrulewidth}
        \emph{k}-NN-acc&  42.5& 44.8 & 47.1 & 48.0 & \textbf{51.4} & 48.5\\
\end{tabular}
\caption{Effect of $\xi$ in the quality of the features learned by the backbone in terms of \emph{k}-NN accuracy with $k=200$, when pre-trained for 100 epochs.}
\label{tab:ablation_epsilon}
\end{table}
\begin{table}
\centering
\begin{tabular}{lccccc}
     $K$& 512  & 1024 &  2048  &  4096  &  8192\\
         \Xhline{2\arrayrulewidth}
        \emph{k}-NN-acc&  44.2& 46.3 & 48.5 & 51.4 & \textbf{52.2}\\
        {queue-memory(MB)}&{4}& {13} & {42} & {148} & {552}\\
\end{tabular}
\caption{Increasing $K$ contributes to the quality of learned features by the model in terms of \emph{k}-NN accuracy with $k=200$, when pre-trained for 100 epochs.}
\label{tab:ablation_K}
\end{table}

\begingroup
\setlength{\tabcolsep}{7pt} 
\renewcommand{\arraystretch}{1.01} 
\begin{table}[htb!]
\centering
\begin{tabular}{ccc}
     Uniform Dist. & Cross-Similarity-Cons. & \emph{k}-NN acc.\\
         \hline
                   \cmark&\cmark&\textbf{51.4}\\
                   \cmark&\xmark&43.8\\
                    \xmark &\cmark&48.5\\
                    \xmark &\xmark&48.5 
\end{tabular}
\caption{Effect of uniform distribution of similarity probability over the negative peers, and cross similarity consistency on the quality of the features learned by the backbone.}
\label{tab:ablation_loss}
\end{table}
\endgroup

\begin{table}
\centering
\begin{tabular}{lcccc}
     {backbone}&{Resnet-50}& {Resnet152}& {Resnet-50xw2}&{Resnet-50xw4} \\
         \Xhline{2\arrayrulewidth}
        {params}& {25m}&{60m}&{98m}&{380m}\\ 
        {top1-acc}&  {71.3}& {73.4} &{74.3} &{75.5}\\
\end{tabular}
\caption{{Effect of backbone on the top-1 accuracy for the linear head protocol on frozen backbone. All the networks are trained for 200 epochs.}}
\label{tab:ablation_backbone}
\end{table}

\begin{table*}[bth!]
\centering
\begin{tabular}{l|c|c|c|c}
  & VOC 07 detection & VOC 07+12 detection  & COCO detection & COCO segmentation\\
\hline
pre-train&  
\begin{tabular}{ccc} $\textrm{AP}$ & $\textrm{AP}_{50}$ & $\textrm{AP}_{75}$ \end{tabular} &
\begin{tabular}{ccc} $\textrm{AP}$ & $\textrm{AP}_{50}$ & $\textrm{AP}_{75}$ \end{tabular}&
\begin{tabular}{ccc} $\textrm{AP}$ & $\textrm{AP}_{50}$ & $\textrm{AP}_{75}$ \end{tabular}&
\begin{tabular}{ccc} $\textrm{AP}^{\textrm{mk}}$ & $\textrm{AP}_{50}^{\textrm{mk}}$ & $\textrm{AP}_{75}^{\textrm{mk}}$ \end{tabular}
\\
\hline
scratch&  
\begin{tabular}{ccc} 16.8 & 35.9 &  13.0 \end{tabular}&
\begin{tabular}{ccc} 33.8 & 60.2 &  33.1 \end{tabular}&
\begin{tabular}{ccc} 26.4 & 44.0 & 27.8 \end{tabular}&
\begin{tabular}{ccc} ~29.3~ &~46.9~ & ~30.8~ \end{tabular}\\
Imagenet-1K-sup.&  
\begin{tabular}{ccc} 42.4& 74.4&  42.7 \end{tabular}&
\begin{tabular}{ccc} 53.5& 81.3&  58.8 \end{tabular}&
\begin{tabular}{ccc} 38.2& 58.2&  41.2 \end{tabular}&
\begin{tabular}{ccc} ~33.3~ &~54.7~ & ~35.2~ \end{tabular}\\
\hline
BYOL\cite{grill2020bootstrap}&  
\begin{tabular}{ccc} 47.1& \textbf{77.3}&  49.7  \end{tabular}&
\begin{tabular}{ccc} 55.4& 81.2&  61.2 \end{tabular}&
\begin{tabular}{ccc} 37.8& 57.7&  40.9 \end{tabular}&
\begin{tabular}{ccc} ~33.1~ &~54.4~ & ~35.1~ \end{tabular}\\
MoCo-v2+\cite{chen2020simple}&
\begin{tabular}{ccc} 48.5& 77.1&  52.5 \end{tabular}&
\begin{tabular}{ccc} 57.0& 82.4&  63.6 \end{tabular}&
\begin{tabular}{ccc} 39.3& 58.7&  42.4 \end{tabular}&
\begin{tabular}{ccc} ~34.2~ &~55.3~ & ~36.5~ \end{tabular}
\\
SwAV*\cite{caron2020unsupervised}&  
\begin{tabular}{ccc}  39.2 & 72.9 & 37.4 \end{tabular}&
\begin{tabular}{ccc} 56.1 & \textbf{82.6} &  62.7 \end{tabular}&
\begin{tabular}{ccc} 38.4 & 58.6 &  41.3 \end{tabular}&
\begin{tabular}{ccc} ~33.8~ &~55.2~ & ~35.9~ \end{tabular}\\
SimSiam\cite{chen2020simple}&
\begin{tabular}{ccc} 48.5& \textbf{77.3}&  52.5  \end{tabular}&
\begin{tabular}{ccc} 57.0& 82.4&  63.7 \end{tabular}&
\begin{tabular}{ccc} 39.2& \textbf{59.3}&  42.1  \end{tabular}&
\begin{tabular}{ccc} ~\textbf{34.4}~ &~\textbf{56.0}~ & ~36.7~ \end{tabular}\\
BarlowT\cite{zbontar2021barlow}&  
\begin{tabular}{ccc} ~--- & ~~~---~ & ~~---~ \end{tabular}&
\begin{tabular}{ccc} 56.8 & \textbf{82.6}&  63.4 \end{tabular}&
\begin{tabular}{ccc} 39.2&  59.0&  42.5 \end{tabular}&
\begin{tabular}{ccc} ~34.3~ &~56.0~ & ~36.5~ \end{tabular}
\\
{VCRReg*\cite{bardes2022vicreg}}&  
\begin{tabular}{ccc} {~---} & {~~~---~} &{~~---~} \end{tabular}&
\begin{tabular}{ccc} {56.7} & {82.4}& {63.3} \end{tabular}&
\begin{tabular}{ccc} {39.1}& { 58.5}&  {42.2} \end{tabular}&
\begin{tabular}{ccc} {~34.1~} &{~56.0~} & {~36.3}\end{tabular}
\\
\hline
XMoCo(ours)& 
\begin{tabular}{ccc} \textbf{48.7} & \textbf{77.3} &  \textbf{52.6} \end{tabular}&
\begin{tabular}{ccc} \textbf{57.1} & 82.5 &  \textbf{63.8} \end{tabular}& 
\begin{tabular}{ccc} \textbf{39.4}&59.1&\textbf{43.0} \end{tabular}&
\begin{tabular}{ccc} ~\textbf{34.4}~ &~55.8~ & ~\textbf{36.9}~ \end{tabular}\\
\end{tabular}
\caption{\footnotesize{Transfer Learning. VOC 07 detection: Faster
R-CNN fine-tuned with VOC 2007 trainval, evaluated on VOC 2007 test; VOC 07+12 detection: Faster R-CNN fine-tuned with VOC 2007
trainval + 2012 train, evaluated on VOC 2007 test.The detector backbone is R50-C4. COCO object detection and instance segmentation: Mask R-CNN-C4 with 1x schedule, trained on COCO2017.}}
\label{tab:table_detection_pascal-R50-C4}
\end{table*}
\subsection{Unsupervised Pre-Training}
Full details of augmentations, optimizer, scheduler and hyper-parameter fine-tuning can be found in  Sec. \ref{sec:unsupervised_training} in the Appendix.
\subsection{Liner head protocol vs end-to-end fine-tuning}We evaluate the quality of the learned visual semantics during self-supervised pre-training, in terms of top-1 accuracy performance. For a fair comparison we have compared all the algorithms when they are pre-trained with two 224$\times$224 views of input images. Our algorithm secures a spot in the top SOTA algorithms in terms of top-1 classification accuracy. We consider two scenarios:\\ 
\textbf{Freezing the backbone weights} and training a linear classifier head for 100 epochs. TABLE~\ref{tab:linear_head_cls} shows the performance of our proposed algorithm compared to existing methods.\\
{\textbf{Fine-tuning the end-to-end network} by using two separate learning rates for the backbone and the linear head, \ie .01 and .2, respectively. We train the network end-to-end for 20 epochs, where we use a \texttt{SGD} optimizer along with a multi-step learning rate scheduler decayed at [12, 16] epoch marks with a decay factor of 0.2. The last column of Table.~\ref{tab:linear_head_cls} shows the performance of fine-tuning the network over Imagenet-1K.}

\subsubsection{Ablation on the number of MLP layers}
We study the effect of number of MLP layers in the quality of the learned visual semantics by ablating the number of layers in TABLE~\ref{tab:effect_mlp_layers}.

Increasing number of MLP layers up to 4, boosts the classification performance. This
is likely because by deepening the model, the last layer of the backbone captures more general low level features, leaving the MLP layers to capture fine-grained pretext task-related features \cite{jing2020self}. It is worth mentioning that even with 3 MLP layers our method performs competitive with other SOTA algorithms. {It must be noted that increasing the number of MLP layers may or may not add to the performance of other SOTA methods; \eg the authors in \cite{zbontar2021barlow} did not achieve any performance improvement when increasing the number of MLP layers for BYOL method. The authors in \cite{chen2020simple} could only reach to better performance for SwAV \cite{caron2020unsupervised} and SimCLR \cite{chen2020simple} when increasing the number of MLP layers by one. Our attempt to gain better top-1 accuracy by increasing the MLP layers on the other SOTA methods failed as depicted in Table.~\ref{tab:mlp_other_sota}.}

\subsubsection{Ablation on the value of $\xi$}
To verify our proposed uniform similarity distribution on the negative peers we ablate the choice of $\xi$ through \emph{k}-nearest-neighbors (\emph{k}-NN) algorithm with $k=200$ neighbors. TABLE~\ref{tab:ablation_epsilon} corroborates our intuition of incorporating the information concealed in the negative peers to the loss via $\xi$. As the value of $\xi$ increases the model learns higher quality features as the similarity between the representations associated to the positive samples increases. However, at the point $\xi = .9$, the trade-off between the positive pair similarity and the interaction between the negative samples reach to an equilibrium, prompting a model learned the highest quality features in terms of \emph{k}-NN accuracy. Increasing $\xi$ towards 1 in practice ignores the role of negative samples and as could be predicted leads to deteriorated quality of learned features.  
\subsubsection{Ablation on the number of negative peers, $K$}
Ablation study as shown in TABLE~\ref{tab:ablation_K} shows that increasing the number of negative peers contributes to the quality of learned features via contrasting against more samples. Additionally, since we use a queue of previous probabilities to deal with the cases where $N\leq K$, larger number of negative samples is neither memory friendly nor is the performance increase worth the hassle. It  must be noted that we have used far less negative pairs, $K=4096$ compared to traditional contrastive methods with $K=65536$ \cite{he2019moco,chen2020mocov2}. 
\subsubsection{Ablation on the loss}\label{sec_ablation_loss}
In this section we study the effect of uniform distribution of similarity on the negative peers as well as the cross similarity consistency in the quality of features learned by the model, by ablating the loss accordingly. As discussed in Sec.~\ref{sec_Discussion} the effect of cross similarity can be cancelled by cancelling the last two terms in the cross-entropy loss of \eqref{eq_cross_entropy_net}. On the other hand the effect of uniform distribution of similarity over the negative peers can be cancelled by using the traditional one-hot-encoded pseudo-labels for our contrastive algorithm. Neutralizing the cross similarity regularization, as shown in TABLE~\ref{tab:ablation_loss} aggravates the quality of the learned features significantly {by losing the symmetry of the loss}. This regularization in short, conveys the transformation invariance property of positive pair in the traditional contrastive learning to negative pairs {if their similarity is non-zero}. Cancelling the uniform distribution of similarity on the negative peers, on the other hand, is equal to setting $\xi=1$ {(one-hot-encoded pseudo-labels)} which naturally cancels the cross similarity regularization as well by decoupling the interaction of negative samples from pseudo-labels. As it can be seen this also exacerbates the quality of the learned features.  
\subsubsection{Ablation on the backbone} The experiments in Table.~\ref{tab:ablation_backbone} ablates the complexity of the backbone instead of the traditional Resnet-50. As the backbone parameters increase the performance also increases.
\subsection{Object Detection and Segmentation}
We use Detectron2 \cite{wu2019detectron2} framework to perform all object detection tasks by finetuning Faster-RCNN \cite{ren2015faster} structures on the pretrained Resnet-50 backbone. We evaluate the results on PASCAL-VOC and COCO datasets and report the standard $\mathrm{AP}$, $\mathrm{AP_{50}}$, and $\mathrm{AP_{75}}$ for both VOC and COCO, all averaged over $5$ trials. We follow the same procedure as in \cite{he2019moco} to fine-tune the model in detectron2. All the experiments are run on $8$ NVIDIA-V100 GPUs with batch size of $16$. 
\subsubsection{PASCAL-VOC Dataset}\label{sec_detect_full_voc}
We fine-tune all layers of a  Faster R-CNN \cite{ren2015faster} detector with a R50-C4 backbone in two different scenarios. 1. We train on the VOC \texttt{trainval07}($\sim 5000$ images) and test on the VOC \texttt{test2007}. 2. We train on the VOC \texttt{trainval07+12} and evaluate on the VOC \texttt{test2007}. We train both with $1\times$ schedules which is approximately 12 epochs. 
TABLE~\ref{tab:table_detection_pascal-R50-C4} shows the detection results fine-tuned on VOC dataset with R50-C4 backbone, where best results are depicted with bolded face font. It can be seen that the performance of XMoCo in localization/classification tasks is either comparable or better than state-of-the-art representation learning methods.
\subsubsection{COCO Object Detection and Segmentation}\label{sec_detect_full_COCO}
Similar to \cite{chen2020exploring} we use a Mask-RCNN \cite{he2017mask} with C4 backbone. We fine tune all layers on \texttt{train2017} and validate on \texttt{val2017} sets. Our proposed XMoCo method performs competitively  with other state-of-the-art methods in COCO object detection and segmentation downstream task.

\section{Appendix}
\begin{figure*}[htb!]
\centering
\centering
\includegraphics[width=.9\textwidth]{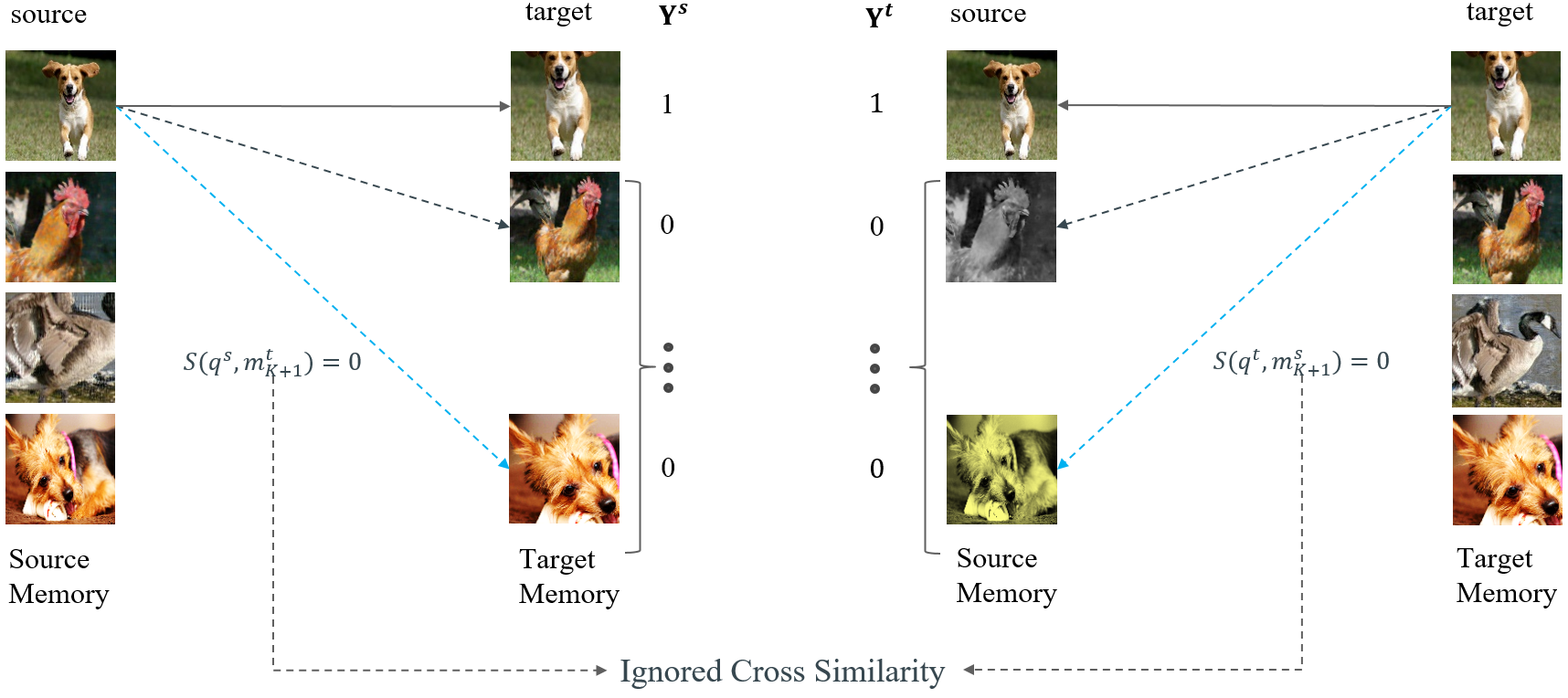}
\caption{Traditional contrastive learning creates a one-hot-encoded labels for the samples in the batch. It ignores the interplay with the negative pairs. As depicted in this figure the two dogs are considered as negative samples, however they have non-zero similarity in reality. Their similarity is forced to be zero in traditional contrastive learning methods.}
\label{fig:con_trad}
\end{figure*}
\begin{figure*}[htb!]
\centering
\includegraphics[width=.9\textwidth]{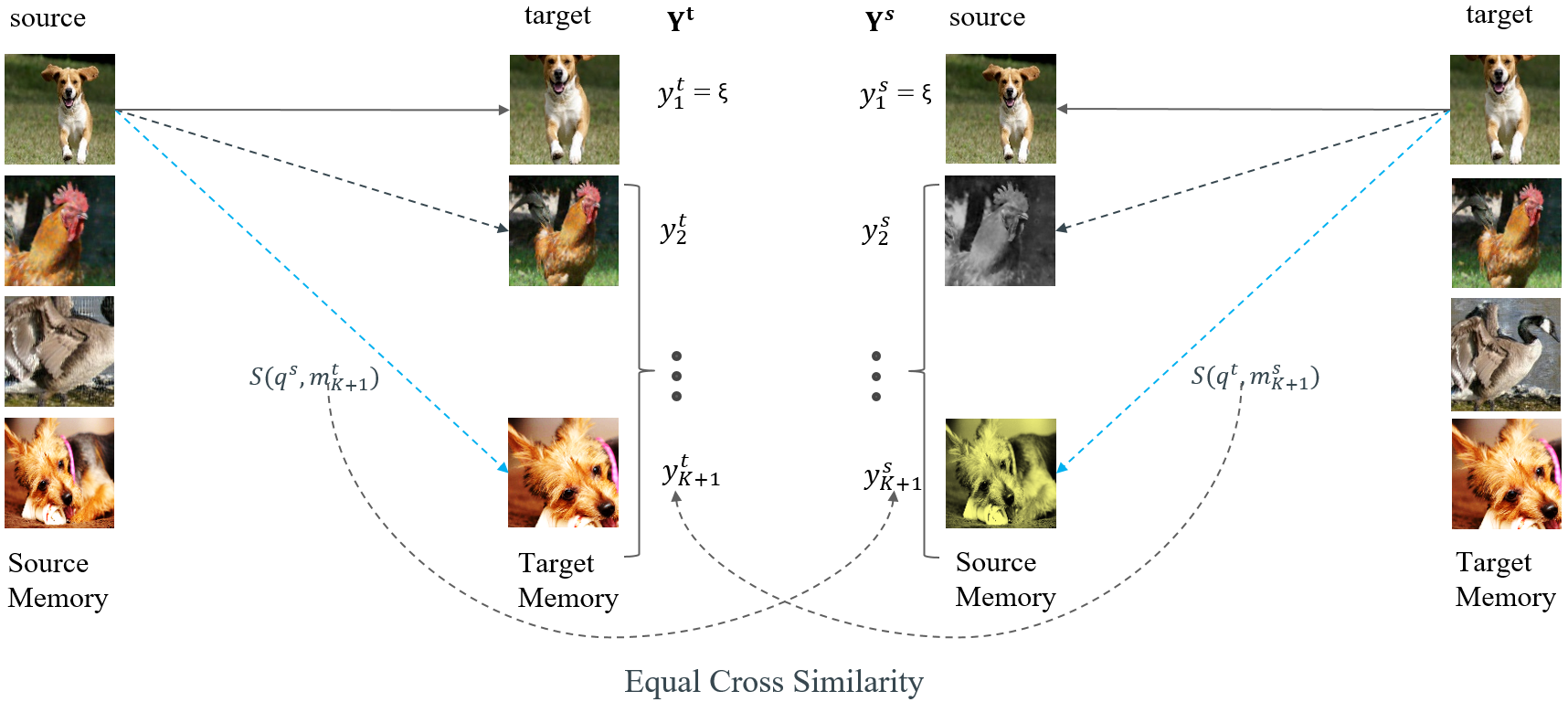}
\caption{XMoCo considers the cross-similarities between the negative samples. By cross-similarity-consistency regularization, XMoCo force $S(\bq^s,\bm^t_{k}) = S(\bq^t,\bm^s_{k})$. Please also note that in XMoCo the labels are swapped. Therefore, the cross similarity between $\bq^s$ and $\bm^t_k$ is forced to follow $S(\bq^t, \bm^s_k)$. Blue lines indicate similarity between the negative samples.}
\label{fig:con_XMoCo}
\end{figure*}

\begin{figure*}[htb!]
\centering
\includegraphics[width=.9\textwidth]{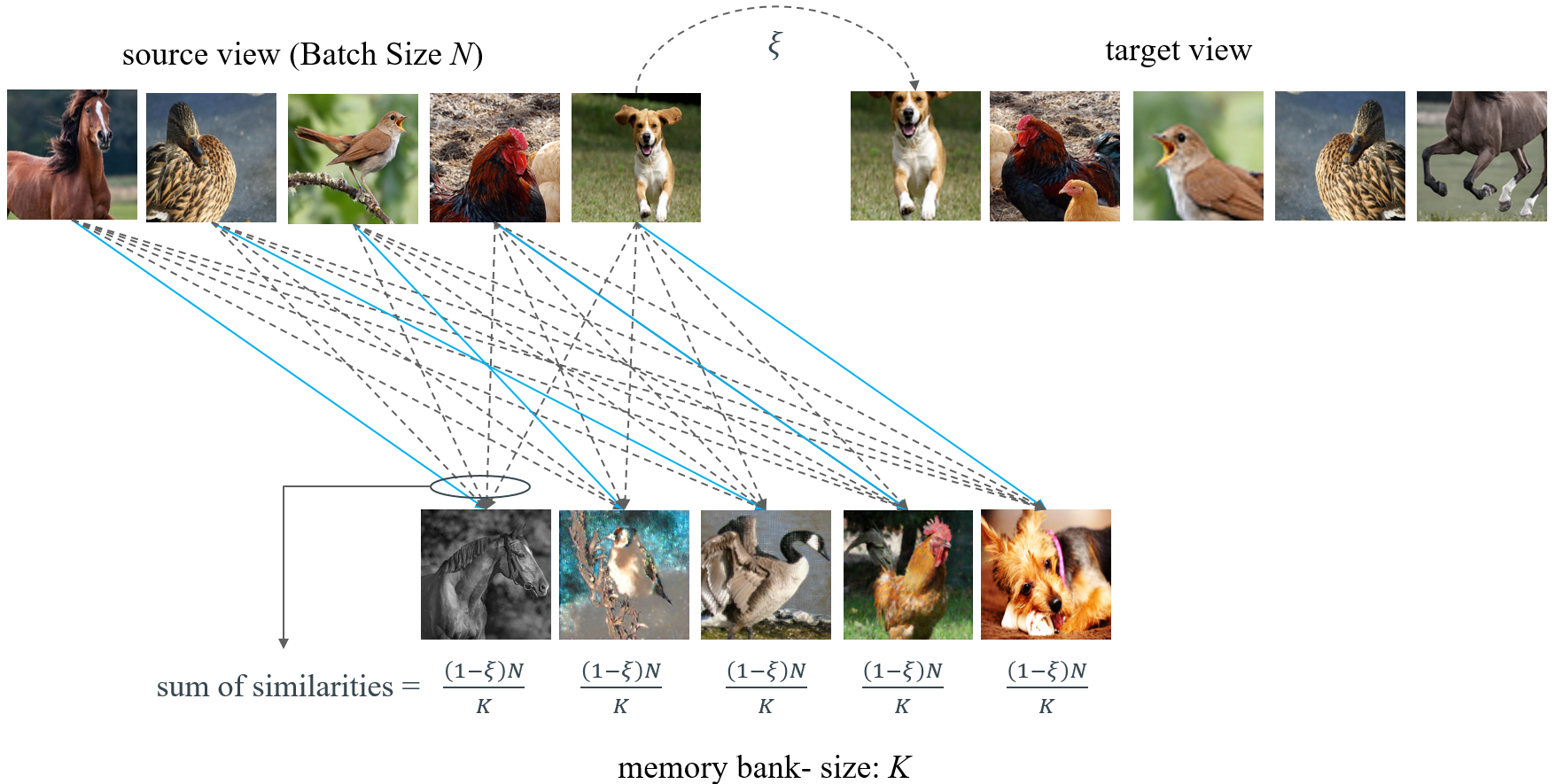}
\caption{Uniform distribution of similarity over the negative peers. For long enough batch of data, each negative sample is equally similar to the whole batch. In other words the aggregated similarity probability for each negative peer is the same. Blue lines show the most similar pairs.}
\label{fig:uniform_dist}
\end{figure*}

\subsection{Traditional Contrastive vs XMoCo}\label{sub:Trad_cons_xmoco}
In traditional contrastive learning the interplay between the negative examples is ignored. Figure \ref{fig:con_trad} shows the schematic of traditional contrastive learning algorithm. As it is shown in the figure, the negative pairs ``\emph{dog-dog}'' are treated similarly as the negative pairs ``\emph{dog-cock}''. In particular all the negative peers in the memory are treated the same. This would degrade the performance of contrastive learning as it pushes away the representation vectors associated to the two dogs as well. The traditional contrastive learning algorithm uses one-hot-encoded pseudo-labels for the probability distributions in \eqref{eq_cne_I} and \eqref{eq_cne_II}. The zeros in the pseudo-labels tend to force the similarity between the negative samples to be zero.

In XMoCo we relax the scenario and let the pseudo-labels to be soft labels. This would allow the model to capture the similarities between \eg the ``\emph{dog-dog}'' negative samples. Figure \ref{fig:con_XMoCo} shows the schematic of the proposed XMoCo algorithm. The pseudo-labels are however constrained to follow the optimization rule in \eqref{eq_max_entropy}, with which we assure that the positive samples still capture the most similarity between all the samples. 
\subsubsection{Cross Similarity Consistency} By swapping the pseudo-labels according to \eqref{eq_cross_entropy_net} we implicitly push the probabilities in \eqref{eq_cne_I} and \eqref{eq_cne_II} to be equal for any \emph{query}-\emph{key} pair which in turn implies \eqref{eq_cross_sim}. Intuitively, in Figure \ref{fig:con_XMoCo} the ``\emph{dog-cock}'' cross similarity in the left is equal to the ``\emph{dog-cock}'' similarity in the right. Note that ``\emph{dog-cock}'' pairs in the left and right are subjected to different \emph{transformations}. 
\subsubsection{Uniform Distribution over the Negative Peers} Figure \ref{fig:uniform_dist} depicts the concept of uniform distribution of similarities over the negative pairs. We can better explain this concept using its analogy to clustering of samples to certain number of clusters. In a classification problem with a balanced dataset, We would like each cluster of data to receive equal number of samples. Here equivalently, we would like each negative sample to receive same aggregated amount of similarity (probability of two representation vectors being the same).
This would regularize out the loss and spreads the similarity of a batch equally to all negative samples.
\subsection{Unsupervised Training}\label{sec:unsupervised_training}
We use \texttt{RandomResizedCrop} in Pytorch, to crop and resize the the input image to two 224 x 224 crops. After that we use \texttt{RandomHorizontalFlip} with probability of 0.5. We also use a random color distortion composed of random \texttt{ColorJitter} with probability $80\%$ and (brightness=0.8, contrast=0.8, saturation=0.8, hue=0.2) and strength of $s=1$, followed by a \texttt{RandomGrayscale} with probability of $0.2$, and a random Gaussian blur with $50\%$ probability and kernel size $23\times 23$ and uniform kernel with unit mean and variance of $0.3$. We normalize the images at the last stage with mean = [0.485, 0.456, 0.406] and  std = [0.228, 0.224, 0.225]. We train the networks using \texttt{SGD} optimizer with momentum of 0.9 and weight decay of 0.0001. We train on 8 Nvidia-V100 GPUs with batch size of 512 and initial learning rate of .135. Similar to \cite{chen2020simple} we use a cosine learning rate schedule \cite{loshchilov2016sgdr} with coefficient of 0.5 and offset of 0.1. 
\subsection{Classification on Imagenet-1K}\label{sec:classification_imagenet}
\subsubsection{1. Linear Head Classification on Imagenet-1K}\label{sub:linear_head}
We further freeze the weights of the encoder that were trained previously in the first stage, and only train the fully connected layer with the true Imagenet-1K labels in a supervised fashion for 100 epochs. The philosophy behind linear head training is that if in the first stage, the image semantics are learned properly, then the linear fully connected layer should learn to classify the images of the same dataset in the second stage within a few epochs. We use Pytorch \texttt{CrossEntropyLoss} and \texttt{SGD} optimizer with momentum of $0.9$ and weight decay of $0.0001$ in this stage. We use batch size of 512 with initial learning rate of 60. We schedule the learning rate to drop by a factor of 0.1 in the interval of [60 80] epochs \cite{wang2015unsupervised}. A softmax layer is concealed in the \texttt{CrossEntropyLoss} function in pytorch.  
\subsubsection{Ablation on the batch size}
TABLE~\ref{tab:ablation_batch_size} shows the effect of changing the batch size on the linear head classification performance, when the backbone is frozen. As observed, the best performance is when the batch size is set to 512.
\begingroup
\setlength{\tabcolsep}{6pt} 
\renewcommand{\arraystretch}{1.01} 
\begin{table}[htb!]
\centering
\begin{tabular}{cccccc}
     batch size  & 256 & 512 & 1024 & 2048 & 4096\\
         \hline
                   top-1 acc & 65.0& \textbf{71.3}& 70.2& 69.5& 58.9\\
                   top-5 acc & 85.3   & \textbf{90.0} & 89.2& 88.7& 81.1\\
\end{tabular}
\caption{Effect of batch size on the linear head classification performance of XMoCo when a Renset-50 backbone is pre-trained for 200 epochs and $K=4096$.}
\label{tab:ablation_batch_size}
\end{table}

\begin{table*}
\centering
\begin{tabular}{l|c|c|c|c|c|c|c|c|c|c}
 pre-train & Aircraft & Caltech101 & Cars & CIFAR10 & CIFAR100 & DTD & Flowers  & Pets & SUN397 & VOC2007\\
\hline
Supervised&43.6&90.2&44.9&91.4&73.9&72.2&89.9&91.5&60.5&83.6 \\
\hline
InsDis\cite{Wu_2018_CVPR}&36.9&71.1&29.0&80.3&60.0&68.5&83.4&68.8&49.5&74.4 \\
MoCo-v1\cite{he2019moco}&35.5&75.3&28.0&80.2&57.7&68.8&82.1&69.8&51.0&75.9 \\
PCL-v1\cite{li2020prototypical}&21.6&76.9&12.9&81.8&55.7&62.9&64.7&75.3&45.7&78.3 \\
PIRL\cite{Misra_2020_CVPR}&37.1&74.5&28.7&82.5&61.3&69.0&83.6&71.4&53.9&76.6 \\
PCL-v2\cite{li2020prototypical}&37.0&86.4&30.5&91.9&73.5&70.6&85.3&82.8&56.2&81.1 \\
SimCLR-v1\cite{chen2020mocov2}&44.9&90.0&43.7&91.2&72.7&74.2&90.9&83.3&59.2&80.8 \\
MoCo-v2\cite{chen2020mocov2}&41.8&87.9&39.3&92.3&74.9&73.9&90.1&83.3&60.3&82.7 \\
SimCLR-v2\cite{chen2020big}&46.4&89.6&50.4&92.5&76.8&76.4&92.9&84.7&61.5&81.6 \\
SeLa-v2\cite{caron2020unsupervised}&37.3&87.2&36.9&92.7&74.8&74.2&90.2&83.2&62.7&82.7 \\
InfoMin\cite{tian2020makes}&38.6&87.8&41.0&91.5&73.4&74.7&87.2&86.2&61.0&83.2 \\
BYOL\cite{grill2020bootstrap}&53.9&\itabold{91.5}&56.4&93.3&77.9&76.9&94.5&89.1&60.0&81.1 \\
DeepCluster-v2\cite{caron2020unsupervised}&54.5&91.3&58.6&\itabold{94.0}&\itabold{79.6}&\textbf{78.6}&94.7&\itabold{89.4}&\itabold{65.5}&\textbf{83.9} \\
SwAV\cite{caron2020unsupervised}&54.0&90.8&54.1&\itabold{94.0}&\itabold{79.6}&77.0&94.6&87.6&\textbf{65.6}&83.7 \\
 {JCL\cite{cai2020joint}}& {52.0}& {89.4}& {53.2}& {93.3}& {77.1}& {74.5}& {93.2}& {86.2}& {63.2}& {81.1}\\
 {SwAV+UTOA*\cite{wang2021improving}}& {55.1}& {90.9}& {55.0}& {94.0}& {78.6}& {78.4}& {95.4}& {87.5}& {65.1}& {81.8}\\

BarlowT\cite{zbontar2021barlow}&\itabold{57.0}&91.0&\textbf{65.0}&92.9&77.8&\itabold{78.5}&\textbf{95.9}&89.3&63.0&82.7\\
\hline
XMoCo(ours)&\textbf{62.9}&\textbf{92.4}&\itabold{64.1}&\textbf{94.9 }&\textbf{80.5}&76.1&\itabold{95.1}&\textbf{89.5}&63.0&\itabold{83.8}
\end{tabular}
   \caption{{Linear transferring of a Resnet-50 model with a pre-trained frozen backbone using logistic regression. For Cifar-10, Cifar100, SUN397, Stanford Cars and DTD the top-1 accuracy is reported. For Pascal VOC 2007 mAP is reported, and for the rest of the datasets, mean accuracy per class. A logistic regression with a cross-entropy loss, and $l_2$ regularization, is trained on the features extracted by the Resnet-50 frozen backbone. We have also reported the result of linear evaluation of Imagent-1K for completeness. The best performance per dataset is in bold font. It must be noted that SwAV, its derivative (UTOA) and DeepCluster-v2 are pre-trained with more than two 224x224 views. The best results are in bold faced font and the second best results are underlined.}}\label{tab:frozen_logistic_regression}
\end{table*}

\begin{table*}
\centering
\centering
\begin{tabular}{l|c|c|c|c|c|c|c|c|c|c}
 pre-train & Aircraft & Caltech101 & Cars & CIFAR10 & CIFAR100 & DTD & Flowers & Pets & SUN397 & VOC2007\\
\hline
Supervised&83.5&91.0&82.6&96.4&82.9&73.3&95.5&92.4&63.6&84.8\\
\hline
InsDis\cite{Wu_2018_CVPR}&73.4&72.0&61.6&93.3&68.3&64.0&89.5&76.2&51.8&71.9 \\
MoCo-v1\cite{he2019moco}&75.6&75.0&65.0&93.9&71.5&65.4&89.5&77.0&53.4&74.9 \\
PCL-v1\cite{li2020prototypical}&75.0&87.6&73.2&96.3&79.6&70.0&90.8&87.0&58.4&82.1 \\
PIRL\cite{Misra_2020_CVPR}&72.7&70.8&61.0&92.2&66.5&64.3&89.8&76.3&50.4&69.9 \\
PCL-v2\cite{li2020prototypical}&79.4&88.0&71.7&96.5&80.3&71.8&93.0&85.4&58.8&82.2 \\
SimCLR-v1&81.1&90.3&83.8&97.1&84.5&71.5&93.8&84.1&63.3&82.6 \\
MoCo-v2\cite{chen2020mocov2}&79.9&84.4&75.2&96.5&71.3&69.5&94.3&79.8&55.8&71.7 \\
SimCLR-v2\cite{chen2020big}&78.7&82.9&79.8&96.2&79.0&70.2&94.3&83.2&61.1&78.2 \\
SeLa-v2\cite{caron2020unsupervised}&82.0&89.0&85.6&96.8&84.4&74.4&95.8&88.5&65.8&\itabold{84.8} \\
InfoMin\cite{tian2020makes}&80.2&83.9&78.8&96.9&71.2&71.1&95.2&85.3&57.7&76.6 \\
BYOL\cite{grill2020bootstrap}&79.5&89.4&84.6&97.0&84.0&73.6&94.5&89.6&64.0&82.7 \\
DeepCluster-v2\cite{caron2020unsupervised}&82.5&90.8&\itabold{87.3}&\itabold{97.1}&\itabold{85.2}&74.8&95.3&89.4&\textbf{66.4}&\textbf{84.9} \\
SwAV\cite{caron2020unsupervised}&83.1&89.8&86.8&96.8&84.4&75.2&95.5&89.0&\itabold{66.2}&84.7 \\
 {JCL\cite{cai2020joint}}& {82.0}& {87.8}& {85.6}& {95.4}& {84.0}& {74.1}& {93.3}& {87.1}& {65.2}& {83.2}\\
 {SwAV+UTOA*\cite{wang2021improving}}& {83.0}& {89.6}& {86.9}& {96.7}& {84.6}& {75.4}& {95.4}& {89.2}& {66.1}& {84.8}\\
BarlowT\cite{zbontar2021barlow}&\itabold{83.9}&\itabold{91.7}&\textbf{88.3}&\itabold{97.1}&84.6&\textbf{76.6}&\itabold{96.6}&\itabold{90.9}&65.5&82.8 \\
\hline
XMoCo(ours)&\textbf{85.4}&\textbf{94.4}&87.0&\textbf{97.5}&\textbf{88.4}&\itabold{76.4}&\textbf{96.8}    &\textbf{93.2}&65.0&\itabold{84.8}
\end{tabular}
   \caption{{Transfer learning of a pre-trained Resnet-50 model by end to end finetuning. We train the backbone and the linear head with two different learning rates. The loss for Pascal VOC 2007 dataset is a binary cross-entropy where as for other datasets we use softmax cross-entropy.  The best results are in bold faced font and the second best results are underlined. We insist that SwAV, its derivative (UTOA) and deepcluster-v2 are trained with 16 image views instead of only 2 views.}}\label{tab:classification_end_to_end}
\end{table*}

\begin{table*}
\begin{subtable}{1\textwidth}
\sisetup{table-format=-1.2}   
\centering
\begin{tabular}{l|c|c|c}
  & VOC 07+12 R50-FPN detection & COCO R50-FPN detection& COCO R50-FPN segmentation\\
\hline
pre-train&  
\begin{tabular}{ccc} $~~\textrm{AP}~~$ & $~~\textrm{AP}_{50}~~$ & $~~\textrm{AP}_{75}~~$ \end{tabular}&
\begin{tabular}{ccc} $~~\textrm{AP}~~$ & $~~\textrm{AP}_{50}~~$ & $~~\textrm{AP}_{75}~~$ \end{tabular}&
\begin{tabular}{ccc} $~~\textrm{AP}^{\textrm{mk}}~~$ & $~~\textrm{AP}_{50}^{\textrm{mk}}~~$ & $~~\textrm{AP}_{75}^{\textrm{mk}}~~$ \end{tabular}
\\
\hline
Imagenet-1K-sup.&  
\begin{tabular}{ccc} $~~51.8~~$&$~~81.0~~$&$~~55.9~~$ \end{tabular}&
\begin{tabular}{ccc} $~~38.5~~$&$~~59.4~~$&$~~41.9~~$ \end{tabular}&
\begin{tabular}{ccc} $35.1~~~~~~$&$56.4~~~$&$~~~37.5~~$ \end{tabular}\\
\hline
XMoCo(ours)&
\begin{tabular}{ccc} $~~\textbf{52.6}~~$&$~~80.2~~$&$~~\textbf{57.9}~~$ \end{tabular}&
\begin{tabular}{ccc} $~~\textbf{39.5}~~$&$~~\textbf{60.1}~~$&$~~\textbf{43.1}~~$ \end{tabular}&
\begin{tabular}{ccc} $\textbf{35.9}~~~~~~$&$\textbf{57.0}~~~$&$~~~\textbf{38.5}~~$ \end{tabular}\\
\end{tabular}
   \caption{{Transfer Learning, effect of backbone: Pascal-VOC object detection and COCO object detection and instance segmentation with R-50-FPN  backbone trained on \texttt{trainval2007+trainval2012} splits, and COCO2017, respectively with $1\times$ schedules.\\\\}}\label{tab:detection_backbone_ablation_first}
\end{subtable}
\begin{subtable}{1\textwidth}
\sisetup{table-format=-1.2}   
\centering
\begin{tabular}{l|c|c|c}
  & VOC 07+12 R50-FPN detection & COCO R50-FPN detection& COCO R50-FPN segmentation\\
\hline
pre-train&  
\begin{tabular}{ccc} $~~\textrm{AP}~~$ & $~~\textrm{AP}_{50}~~$ & $~~\textrm{AP}_{75}~~$ \end{tabular}&
\begin{tabular}{ccc} $~~\textrm{AP}~~$ & $~~\textrm{AP}_{50}~~$ & $~~\textrm{AP}_{75}~~$ \end{tabular}&
\begin{tabular}{ccc} $~~\textrm{AP}^{\textrm{mk}}~~$ & $~~\textrm{AP}_{50}^{\textrm{mk}}~~$ & $~~\textrm{AP}_{75}^{\textrm{mk}}~~$ \end{tabular}
\\
\hline
Imagenet-1K-sup.&
\begin{tabular}{ccc} $~~54.2~~$&$~~81.5~~$&$~~59.7~~$ \end{tabular}&
\begin{tabular}{ccc} $~~40.7~~$&$~~61.2~~$&$~~44.7~~$ \end{tabular}&
\begin{tabular}{ccc} $37.0~~~~~~$&$58.3~~~$&$~~~39.9~~$ \end{tabular}\\
\hline
XMoCo(ours)&   
\begin{tabular}{ccc} $~~\textbf{55.9}~~$&$~~\textbf{82.0}~~$&$~~\textbf{61.7}~~$ \end{tabular}&
\begin{tabular}{ccc} $~~\textbf{42.0}~~$&$~~\textbf{62.8}~~$&$~~\textbf{46.0}~~$ \end{tabular}&
\begin{tabular}{ccc} $\textbf{38.0}~~~~~~$&$\textbf{59.9}~~~$&$~~~\textbf{41.0}~~$ \end{tabular}\\
\end{tabular} 
   \caption{{Transfer Learning, effect of longer training: Pascal-VOC object detection and COCO object detection and instance segmentation with R-50-FPN  backbone trained on \texttt{trainval2007+trainval2012} splits, and COCO2017, respectively with $3\times$ schedules.\\\\}}\label{tab:detection_backbone_ablation_second}
\end{subtable}

\vspace{-16pt}
\caption{Object detection and segmentation ablations on backbone and training length} \label{tab:detection_backbone_ablation}
\end{table*}
\begingroup
\begin{table*}[htb!]
\centering
\begin{tabular}{l|c|c}
 pre-train & COCO R50-FPN-RetinaNet detection, 1x schedule& COCO R50-FPN-RetinaNet detection, 2x schedule\\
\hline
&  
\begin{tabular}{ccc} $\quad\textrm{AP}\quad$ & $\quad\textrm{AP}_{50}\quad$ & $\quad\textrm{AP}_{75}\quad$ \end{tabular}&
\begin{tabular}{ccc} $\quad\textrm{AP}\quad$ & $\quad\textrm{AP}_{50}\quad$ & $\quad\textrm{AP}_{75}\quad$ \end{tabular}\\

\hline
Imagenet-1K-sup.& 
\begin{tabular}{ccc} $\quad37.0\quad$ & $\quad56.1\quad$ & $\quad39.8\quad$ \end{tabular}&
\begin{tabular}{ccc} $\quad39.3\quad$ & $\quad58.8\quad$ & $\quad42.4\quad$ \end{tabular}\\
\hline
XMoCo(ours)&  
\begin{tabular}{ccc} $\quad\textbf{37.6}\quad$ & $\quad\textbf{57.0}\quad$ & $\quad\textbf{40.2}\quad$ \end{tabular}&
\begin{tabular}{ccc} $\quad\textbf{39.8}\quad$ & $\quad\textbf{59.6}\quad$ & $\quad\textbf{42.5}\quad$ \end{tabular}
\end{tabular}
\caption {COCO object detection with R-50-FPN backbone and RetinaNet detector with 1x and 2x schedules.}
\label{tab:detection_retina_net}
\end{table*}

\subsection{Classification on other Datasets}\label{sec:classification_other}
To study how well features learned during self-supervised training can be transferred to other datasets, we study the classification downstream task on various target datasets FGVC Aircraft \cite{maji2013fine}, Caltech-101 \cite{fei2004learning}, Stanford Cars \cite{7299023}, CIFAR10, and  CIFAR-100 \cite{krizhevsky2009learning}, DTD \cite{6909856}, Oxford 102 Flowers\cite{nilsback2008automated}, Oxford-IIIT Pets \cite{parkhi12a}, SUN397 \cite{5539970}
and Pascal VOC2007 \cite{everingham2010pascal}. These datasets span a wide variety of classification tasks such as fine-grained, texture and scene classification. The cardinality of the target dataset and their associated number of classes span the intervals [2000, 75000] and [10, 397], respectively.
For the target datasets Cifar-10, Cifar100, SUN397, Stanford Cars and DTD the top-1 accuracy is reported; however, due to the multi-label nature of datasets like Pascal VOC 2007 the 11 point interpolated average precision, mAP, is reported \cite{everingham2010pascal}. For the rest of the target datasets we report mean accuracy per class. We split the train, validation and test splits for each datasets similar to what it is reported in \cite{ericsson2021selfsupervised}. We further compare our results with respect to the representation learning pioneers. 
\cite{Wu_2018_CVPR,he2019moco,asano2019self,
Misra_2020_CVPR,li2020prototypical,chen2020mocov2,chen2020big,
tian2020makes,grill2020bootstrap,
caron2020unsupervised,zbontar2021barlow}.


We study the classification transfer learning for the scenarios explained in the following two sections.

\subsubsection{Linear Evaluation Using Logistic Regression}
We train a logistic regression with a cross-entropy loss,  a L-BFGS \cite{liu1989limited} solver, and an $l_2$ regularization, on the features extracted by the Resnet-50 frozen backbone. We train for 5000 iterations for a batch size of 64. We use the same augmentation for both train and validation datasets. In particular, we resize the input image to 224 on the small edge and apply a center crop following with a normalization with the Imagenet-1K normalization factors. We fine-tune the regularization factor similar to \cite{ericsson2021selfsupervised}. Since Pascal VOC 2007 is a multi-label dataset, we fit 20 binary logistic regression for each class and report the mAP.

Our method achieves the top-2 spot in 8 datasets out of 10 as depicted in TABLE~\ref{tab:frozen_logistic_regression}. It must be noted that methods like SwAV, DeepCluster-v2, and SeLa-v2 \cite{caron2020unsupervised} are pre-trained on 16 views rather than two 224 x 224 views. Our method performs well in almost all the scenarios, from Aircraft fine-grained classification to Pascal VOC 2007 multi-label classification.
\subsubsection{Transfer Learning (End-to-End Fine-Tuning)}
In this section we append a linear head to the pre-trained Resnet-50 backbone and fine-tune the network end-to-end with a binary cross-entropy loss for Pascal VOC 2007 and softmax cross-entropy loss for all other datasets. We use the same training settings and augmentations as before. Additionally we use a \texttt{SGD} optimizer with nestrov momentum of 0.9 and a \texttt{CosineAnnealingLR} learning rate scheduler for 5000 iterations.  We grid search for the best learning rate between 10 logarithmically equally spaced points the interval [$10^{-4}$, $10^{-1}$]. We also grid search for the best optimizer weight decay among 10 logarithmically equally spaced points in the interval [$10^{-6}$, $10^{-3}$] appended with a zero weight decay point. 

When fine-tuning the pre-trained model on various datasets with the aforementioned specifications, as shown in TABLE~\ref{tab:classification_end_to_end}, our method performs very well and achieves a top spot in the classification accuracy in 8 out of 10 datasets. 
\subsection{Object Detection and Segmentation}\label{sec:object_detection}
Due to limitations of the Batch Normalization (BN) \cite{ioffe2015batch} in structures like Resnet-50 it is relatively hard to achieve good detection results by training object detectors from scratch \cite{he2019rethinking}\footnote{Object detectors are trained on images of higher resolution. This makes it inevitable to load batches of very small size to the GPUS due to memory limitations. Therefore, BN operates on very small batch sizes, resulting in inaccurate learned parameters \cite{ioffe2017batch,8578745,wu2018group}. }. Therefore, it is common to pre-train on Imagenet-1K in a supervised fashion, and then freeze the BN layers and fine-tune the detector on the downstream task by using the pre-trained network as the initialization. Detectron2 as a standard platform however, trains object detectors with selected standard hyper-parameters fine-tuned on Imagenet-1K supervised pre-trained networks. Since, detectors are highly dependant to their training hyper-parameters and the scheduling mechanism, for a fair comparison, we follow the normalization routine adopted by \cite{he2019moco} via fine-tuning and synchronizing the BN trained across GPUs \cite{8578745}. This would bypass the need for freezing the BN layers trained on Imagenet-1K in the detector architecture \cite{he2016deep}. Also, similar to \cite{he2019moco} we normalize the detector-specific appended layers by adding BN for better adjustment of the weights magnitude. As for fine-tuning, we use the same schedule as the Imagenet-1K supervised pre-training peer.    
\subsubsection{Pascal-VOC Object Detection}
We use a Faster R-CNN \cite{ren2015faster} detector with R50-C4 backbone, The image scale during training changes in the interval $[480~800]$ with the step of $32$ pixels, where at the test time it is fixed on $800$ pixels. Unless otherwise stated, we train for $12k$ iterations where we reduce the learning rate by $\times10$ in $8k$ and $10k$ iteration marks. The initial learning rate is set as the default value of $.02$ with linear warm-up \cite{goyal2017accurate} for $100$ iterations. The weight decay and momentum are $0.0001$ and $0.9$, respectively. 
\subsubsection{COCO Object Detection and Segmentation}
Similar to \cite{chen2020exploring} we use a Mask-RCNN \cite{he2017mask} with C4 backbone. We train for $90k$ iterations, in the main experiments, with stepping down the learning rate by $\times 10$ at the iterations $60k$ and $80k$ (The $\times 1$ schedule routine). We fine tune all layers on \texttt{train2017} and validate on \texttt{val2017} sets. The image size for training is within [640, 800] and at the inference is 600. The batch size is 16 and the learning rate is the default value of $.02$.

\subsubsection{Ablation on the backbone}
TABLE~\ref{tab:detection_backbone_ablation_first} investigates the effect of changing the detector backbone on the detection performance compared to the supervised training counterpart. As it can be seen XMoCo outperforms its supervised counterpart when trained on R50-FPN backbone for both VOC 2007 and COCO datasets. The training in this section is executed for 12 epochs only (1x  schedule).  
\subsubsection{Ablation on the training schedules}
Table \ref{tab:detection_backbone_ablation_second} shows how longer schedules can affect the detection performance in both Pascal and COCO datasets when the backbone is R50-C4. 
As it is depicted in the table, not only does XMoCo hold to out performing its supervised counterpart, but also the gap between the unsupervised and supervised performances is increased.

\subsubsection{Ablation on the detector}
In this section instead of a Mask-RCNN \cite{he2017mask} we train a RetinaNet \cite{lin2018focal} detector with R50-FPN backbone on COCO dataset. Also with RetinaNet detector, as depicted in TABLE~\ref{tab:detection_retina_net}, XMoCo outperforms its supervised training rival in terms of average precision.

\section{Conclusion}
\label{sec:conclusion}
Traditional contrastive self-supervised representation learning methods in general pull the representations associated to the positive pairs together and push the ones associated to the negative pairs away. Traditionally, these methods ignore the inter-connection between the negative pairs. In this paper we presented extended momentum contrast, XMoCo, a method founded upon the legacy MoCo contrastive learning family.
we introduced a cross consistency regularization loss, through which we extend the pretext invariance idea presented in the conventional representation learning methods to dissimilar images (negative pairs). Under the cross consistency regularization rule, we argued that transformations should preserve the cross similarities between the negative sample representation vectors as well. Additionally, we also proposed to soften the one-hot encoded pretext pseudo-labels in the classical infoNCE loss to take the cross similarities of the negative pairs into account. We proposed a formulation to uniformly distribute the similarity of samples in a batch, among all the negative peers in the memory bank. 
Our results show a competitive performance on the standard self-supervised learning benchmarks with respect to other SOTA representation learning methods.

{\small
\bibliographystyle{IEEEtran}
\bibliography{IEEEabrv,references}
}

\begin{IEEEbiography}
[{\includegraphics[width=0.9in,height=1.1in,clip]{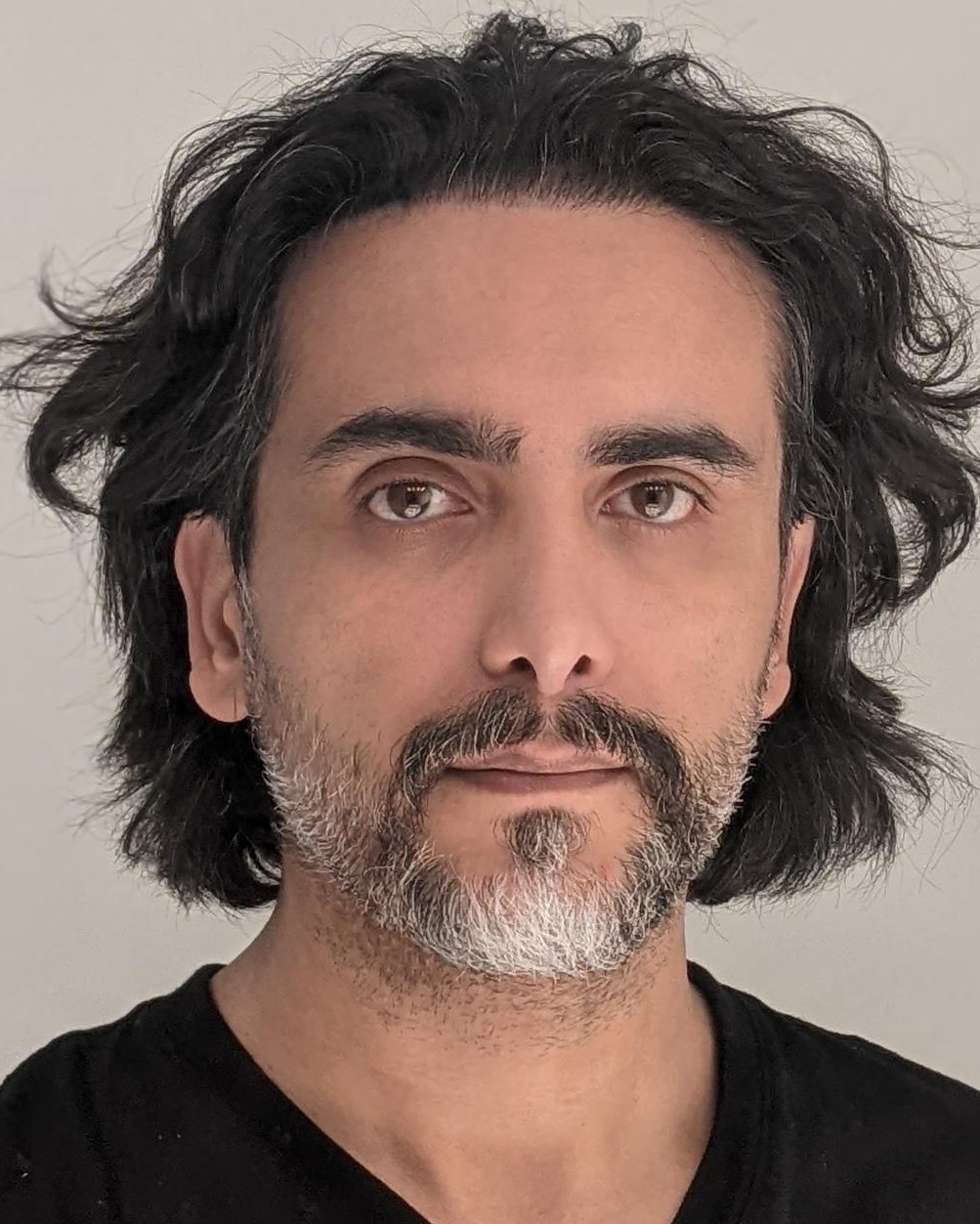}}]{Mehdi Seyfi}
is a staff researcher in the Vancouver big data and intelligence platform lab at Huawei Technologies Canada. His research interest spans various domains in computer vision, image/video/signal processing, machine learning, and deep learning. Mehdi is the recipient of the Simon Fraser University Dean of Graduate Studies (DGS) convocation medal of excellence for his academic achievements during his PhD.
\end{IEEEbiography}
\begin{IEEEbiography}
[{\includegraphics[width=0.9in,height=1.1in,clip]{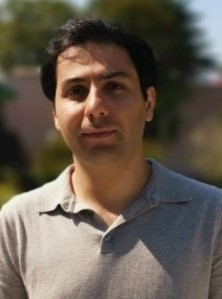}}]{Amin Banitalebi-Dehkordi}
received his PhD from the University of British Columbia (UBC), Canada, in 2014. His academic career has resulted in publications in computer vision and pattern recognition, visual attention modeling, and high dynamic range video. His industrial experience expands to areas in machine learning, deep learning, computer vision, and NLP. Amin is currently a principal researcher in machine learning and technical lead at Huawei Technologies Canada.
\end{IEEEbiography}
\begin{IEEEbiography}
[{\includegraphics[width=0.9in,height=1.1in,clip]{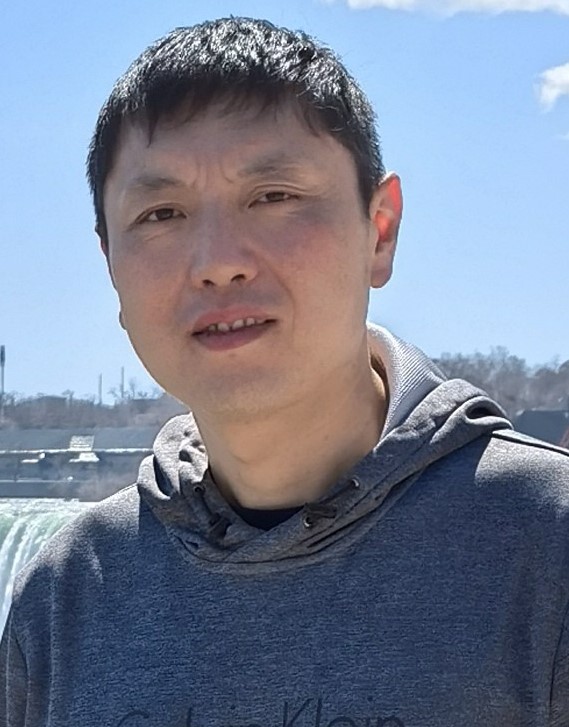}}]{Yong Zhang}
is a Distinguished Researcher at Huawei Technologies Canada and leading the big data and intelligence platform lab at Vancouver research center. Prior to that, he was a postdoctoral research fellow at Stanford University, USA. His research interests include large scale numerical optimization and machine learning. His research works have been published in top-tier journals and conferences.
\end{IEEEbiography}

\end{document}